\documentclass[letterpaper]{article} 
\usepackage{aaai24}  
\usepackage{times}  
\usepackage{helvet}  
\usepackage{courier}  
\usepackage[hyphens]{url}  
\usepackage{graphicx} 
\usepackage{multirow}
\usepackage{hhline}
\usepackage{pifont}
\usepackage{booktabs}
\usepackage{natbib}  
\usepackage{caption} 
\usepackage{algorithm}
\usepackage{algorithmic}
\usepackage{amsmath}
\usepackage{newfloat}
\usepackage{listings}
\DeclareCaptionStyle{ruled}{labelfont=normalfont,labelsep=colon,strut=off} 
\floatstyle{ruled}
\newfloat{listing}{tb}{lst}{}
\floatname{listing}{Listing}
\title{TOP-ReID: Multi-spectral Object Re-Identification with Token Permutation}
\author{
    Yuhao Wang\textsuperscript{\rm 1},
    Xuehu Liu\textsuperscript{\rm 2},
    Pingping Zhang\textsuperscript{\rm 1}\thanks{Corresponding author (zhpp@dlut.edu.cn).},
    Hu Lu\textsuperscript{\rm 3},
    Zhengzheng Tu\textsuperscript{\rm 4},
    Huchuan Lu\textsuperscript{\rm 1}
}
\affiliations{
    \textsuperscript{\rm 1}School of Future Technology, School of Artificial Intelligence, Dalian University of Technology\\
    \textsuperscript{\rm 2}School of Computer Science and Artificial Intelligence, Wuhan University of Technology\\
    \textsuperscript{\rm 3}School of Computer Science and Communication Engineering, Jiangsu University\\
    \textsuperscript{\rm 4}School of Computer Science and Technology, Anhui University\\

}

\usepackage{bibentry}

\begin{document}

\maketitle

\begin{abstract}
Multi-spectral object Re-identification (ReID) aims to retrieve specific objects by leveraging complementary information from different image spectra.
It delivers great advantages over traditional single-spectral ReID in complex visual environment.
However, the significant distribution gap among different image spectra poses great challenges for effective multi-spectral feature representations.
In addition, most of current Transformer-based ReID methods only utilize the global feature of class tokens to achieve the holistic retrieval, ignoring the local discriminative ones.
To address the above issues, we step further to utilize all the tokens of Transformers and propose a cyclic token permutation framework for multi-spectral object ReID, dubbled TOP-ReID.
More specifically, we first deploy a multi-stream deep network based on vision Transformers to preserve distinct information from different image spectra.
Then, we propose a Token Permutation Module (TPM) for cyclic multi-spectral feature aggregation.
It not only facilitates the spatial feature alignment across different image spectra, but also allows the class token of each spectrum to perceive the local details of other spectra.
Meanwhile, we propose a Complementary Reconstruction Module (CRM), which introduces dense token-level reconstruction constraints to reduce the distribution gap across different image spectra.
With the above modules, our proposed framework can generate more discriminative multi-spectral features for robust object ReID.
Extensive experiments on three ReID benchmarks (i.e., RGBNT201, RGBNT100 and MSVR310) verify the effectiveness of our methods.
The code is available at https://github.com/924973292/TOP-ReID.
\end{abstract}
\section{Introduction}
\begin{figure}[t]
    \centering
    \includegraphics[width=0.40\textwidth]{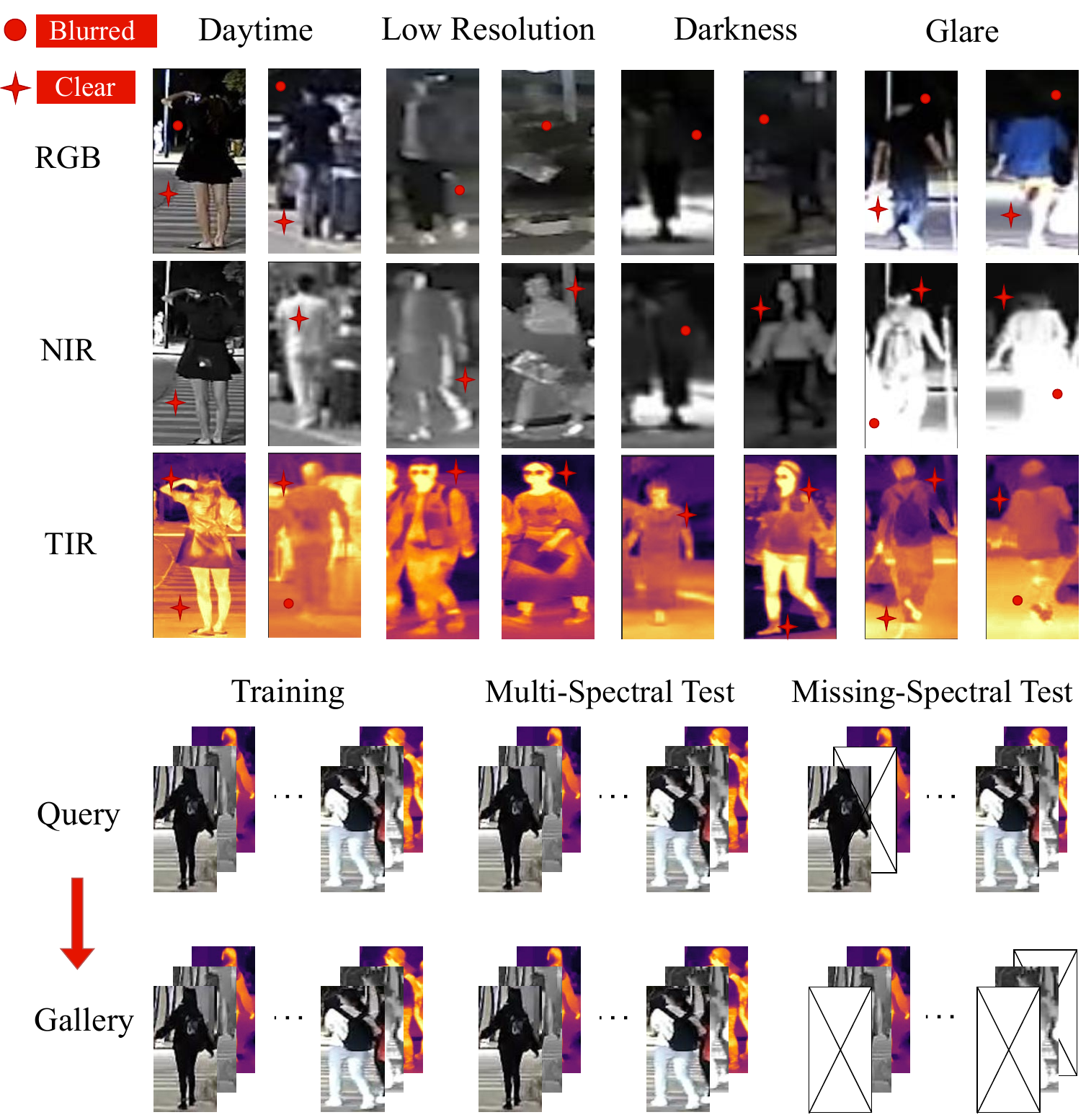}
    \caption{The top displays instances from RGBNT201 in various challenges, while the bottom presents the object ReID settings for multi-spectral and missing-spectral test.}
    \label{fig:Introduction}
\end{figure}
Object Re-identification (ReID) aims to retrieve specific objects from images or videos across non-overlapping cameras, which has advanced significantly over the past decades.
In the traditional object ReID, researchers primarily utilize single-spectral images (such as RGB, depth) to extract visual information of the targets.
However, single-spectral images provide very limited representation abilities in scenarios characterized by low resolution, darkness, glare, etc.
As illustrated in the top row of Fig. \ref{fig:Introduction}, the outlines of persons are notably blurred, leading to an evident confusion between persons and the background in the RGB image spectrum.
Hence, relying only on RGB images poses great challenges for robust object ReID.
Fortunately, other image spectra are very useful to address above problems.
In fact, Near Infrared (NIR) imaging is unaffected by darkness and adverse weather conditions~\cite{li2020multi}.
Thus, there have been some efforts~\cite{li2020infrared,liu2021strong,zhang2023diverse} to incorporate NIR images to enhance the performance of object ReID.
Nonetheless, NIR images retain some limitations \cite{zheng2021robust}, as depicted in Fig. \ref{fig:Introduction}.
For example, the details of persons in NIR images tend to be substantially obscured in the presence of glare.
Meanwhile, Thermal Infrared (TIR) imaging is more robust to these scenarios~\cite{zheng2021robust}.
As illustrated in Fig. \ref{fig:Introduction}, TIR images can highlight persons from the background and preserve crucial details, such as glasses and backpacks.
These facts clearly show the information complementarity of different image spectra for object ReID.
Based on the above facts, multi-spectral object ReID aims to retrieve specific objects by leveraging complementary information from different image spectra, e.g., RGB, NIR, TIR etc.
It delivers great advantages over single-spectral ReID in complex visual environment.
In fact, some methods~\cite{zheng2021robust,wang2022interact} have already tried to integrate multi-spectral features with simple fusion methods.
However, there are significant distribution gaps among different image spectra.
Simple fusions can not well address the heterogenous challenges for effective feature representations.
In addition, it often involves the absence of image spectra in real world, as shown in the ``Missing-spectral Test'' of Fig. \ref{fig:Introduction}.
Thus, there are much room for improving multi-spectral feature fusion.

Meanwhile, with the great advance of vision Transformers~\cite{dosovitskiy2020image}, some works~\cite{he2021transreid,pan2022h} have introduced Transformers for object ReID.
However, most of current Transformer-based ReID methods only utilize the global feature of class tokens to achieve the holistic retrieval, ignoring the local discriminative ones.
To address the above issues, we step further to utilize all the tokens of Transformers and propose a cyclic token permutation framework for multi-spectral object ReID, dubbled TOP-ReID.
Specifically, it consists of two key modules: Token Permutation Module (TPM) and Complementary Reconstruction Module (CRM).
Technically, we first deploy a multi-stream deep network based on vision Transformers to preserve distinct information from different image spectra.
Then, TPM takes all the tokens from the multi-stream deep network as inputs, and cyclically permutes the specific class tokens and the corresponding patch tokens from other spectra.
In this way, it not only facilitates the spatial feature alignment across different image spectra, but also allows the class token of each spectrum to perceive the local details of other spectra.
Meanwhile, CRM is proposed to facilitate local information interaction and reconstruction across different image spectra.
Through introducing token-level reconstruction constraints, it can reduce the distribution gap across different image spectra.
As a result, the CRM can further handle the missing-spectral problem.
With the proposed modules, our framework can extract more discriminative features from multi-spectral images for robust object ReID.
Comprehensive experiments are conducted on three multi-spectral object ReID benchmarks, i.e., RGBNT201, RGBNT100 and MSVR310.
Experimental results clearly show the effectiveness of our proposed methods.

In summary, our contributions can be stated as follows:
\begin{itemize}
\item
We propose a novel feature learning framework named TOP-ReID for multi-spectral object ReID.
To our best knowledge, our proposed TOP-ReID is the first work to utilize all the tokens of vision Transformers to improve the multi-spectral object ReID.
\item
We propose a Token Permutation Module (TPM) and a Complementary Reconstruction Module (CRM) to facilitate multi-spectral feature alignment and handle spectral-missing problems effectively.
\item
We perform comprehensive experiments on three multi-spectral object ReID benchmarks, i.e., RGBNT201, RGBNT100 and MSVR310. The results fully verify the effectiveness of our proposed methods.
\end{itemize}
\section{Related Work}
\subsection{Single-spectral Object ReID}
Single-spectral object ReID focuses on extracting discriminative features from single-spectral images.
Typical single-spectral forms include RGB, NIR, TIR and depth.
Due to the easy requirement, RGB images play a fundamental role in the single-spectral object ReID.
As for the techniques, most of existing object ReID methods are based on Convolutional Neural Networks (CNNs).
%
For example, Luo \emph{et al.}~\cite{luo2019bag} utilize a deep residual network and introduce the BNNeck technique for object ReID.
Furthermore, PCB \cite{sun2018beyond} and MGN \cite{wang2018learning} adapt a stripe-based image division strategy to obtain multi-grained representations.
OSNet \cite{zhou2019omni} employs a unified aggregation gate for fusing omni-scale features.
AGW \cite{ye2021deep} incorporates non-local attention mechanisms for fine-grained feature extraction.
Nevertheless, due to the limited receptive field, CNN-based methods\cite{qian2017multi,li2018harmonious,chang2018multi,chen2019deep,sun2020circle,rao2021counterfactual,zhao2021heterogeneous,Liu_2021_CVPR} are not robust to complex scenarios.
Inspired by the success of vision Transformers (ViT)~\cite{dosovitskiy2020image}, He \emph{et al.}~\cite{he2021transreid} propose the first pure Transformer-based method named TransReID for object ReID, yielding competitive results through the adaptive modeling of image patches.
Afterwards, numerous Transformer-based methods~\cite{zhu2021aaformer,zhang2021hat,chen2022rest,wang2022nformer,liu2023deeply} demonstrate their advantages in object ReID.
However, all these methods take single-spectral images as inputs, providing limited representation abilities.
Thus, they can not handle the all-day object ReID problem.
\begin{figure*}[t]
\centering
\includegraphics[width=0.90\textwidth]{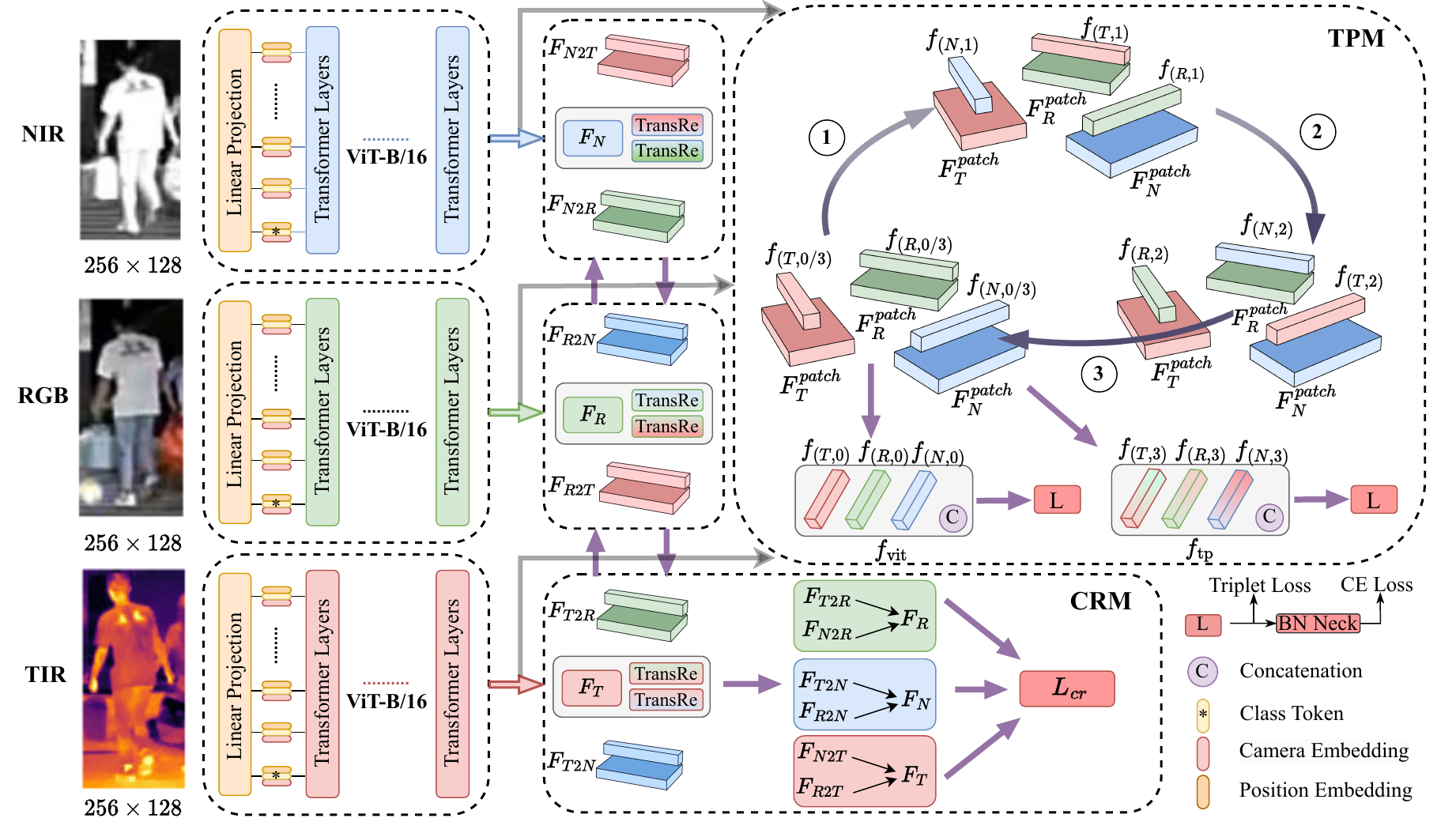}
\caption{An illustration of the proposed TOP-ReID.
First, deep features from RGB, NIR and TIR images are extracted by using three independent ViT-B/16.
Then, a Token Permutation Module (TPM) is proposed for cyclic multi-spectral feature aggregation through three consecutive token permutations.
Meanwhile, a Complementary Reconstruction Module (CRM) is used to achieve token-level reconstruction constraints.
When inference, we utilize the permutated features  for ranking the person candidates.}
\label{fig:Overall}
\end{figure*}
\subsection{Multi-spectral Object ReID}
The robustness of multi-spectral data draws the attention of numerous researchers.
For multi-spectral person ReID, Zheng \emph{et al.}~\cite{zheng2021robust} advance the field and design a PFNet to learn robust RGB-NIR-TIR features.
Then, Wang \emph{et al.}~\cite{wang2022interact} boost modality-specific representations with three learning strategies, named IEEE.
Furthermore, Zheng \emph{et al.}~\cite{zheng2023dynamic} design a DENet to address the spectral-missing problem.
For multi-spectral vehicle ReID, Li \emph{et al.}~\cite{li2020multi} propose a HAMNet to fuse different spectral features.
Considering the relationship between different image spectra, Guo \emph{et al.}~\cite{guo2022generative} propose a GAFNet to fuse the multiple data sources.
He \emph{et al.}~\cite{he2023graph} propose a GPFNet to adaptively fuse multi-spectral features.
Zheng \emph{et al.}~\cite{zheng2022multi} propose a CCNet to simultaneously overcome the discrepancies from both modality and sample aspects.
Pan \emph{et al.}~\cite{pan2022h} propose a HViT to balance modal-specific and modal-shared information.
Furthermore, they employ a random hybrid augmentation and a feature hybrid mechanism to improve the performance~\cite{pan2023progressively}.
Although effective, previous methods mainly treat the NIR and TIR as an assistant to RGB, rather than adaptively fuse them with multi-level spatial correspondences.
In contrast, we facilitate the spatial feature alignment across different image spectra.
\section{Proposed Method}
As illustrated in Fig. \ref{fig:Overall}, our proposed TOP-ReID consists of three main components: Multi-stream Feature Extraction, Token Permutation Module (TPM) and Complementary Reconstruction Module (CRM).
\subsection{Multi-stream Feature Extraction}
In this work, we take images of three spectra for object ReID, i.e., RGB, NIR and TIR.
To capture the distinctive characteristics of each spectrum, we follow previous works~\cite{li2020multi,zheng2021robust} and adopt three independent backbones.
More specifically, vision Transformers (ViT) can be deployed as the backbone in each stream.
Formally, the multi-stream features can be represented as
\begin{equation}
    F_{\mathrm{R}} = \mathrm{ViT}_{\mathrm{R}}\left(I_{\mathrm{R}}\right),
    \end{equation}
\begin{equation}
    F_{\mathrm{N}} = \mathrm{ViT}_{\mathrm{N}}\left(I_{\mathrm{N}}\right),
    \end{equation}
    \begin{equation}
    F_{\mathrm{T}} =  \mathrm{ViT}_{\mathrm{T}}\left(I_{\mathrm{T}}\right),
\end{equation}
where $I_\mathrm{R} \in R^{H \times W \times 3}$, $I_\mathrm{N}\in R^{H \times W \times 3}$ and $I_\mathrm{T}\in R^{H \times W \times 3}$ denote the input RGB, NIR and TIR images, respectively.
Here, $\mathrm{ViT}$ can be any vision Transformers (e.g., ViT-B/16~\cite{dosovitskiy2020image}, DeiT-S/16~\cite{touvron2021training}, T2T-ViT-24~\cite{yuan2021tokens}).
The token features $F_\mathrm{R}$, $F_\mathrm{N}$, $F_\mathrm{T} \in R^{D \times (M+1)}$ are extracted from the final layer of $\mathrm{ViT}$, respectively.
Additional learnable class token is included.
$D$ denotes the embedding dimension while $M$ means the number of patch tokens.
These independent streams enable the extraction of spectral-specific features, capturing rich information from different image spectra.
\begin{figure}[t]
\centering
\includegraphics[width=0.47\textwidth]{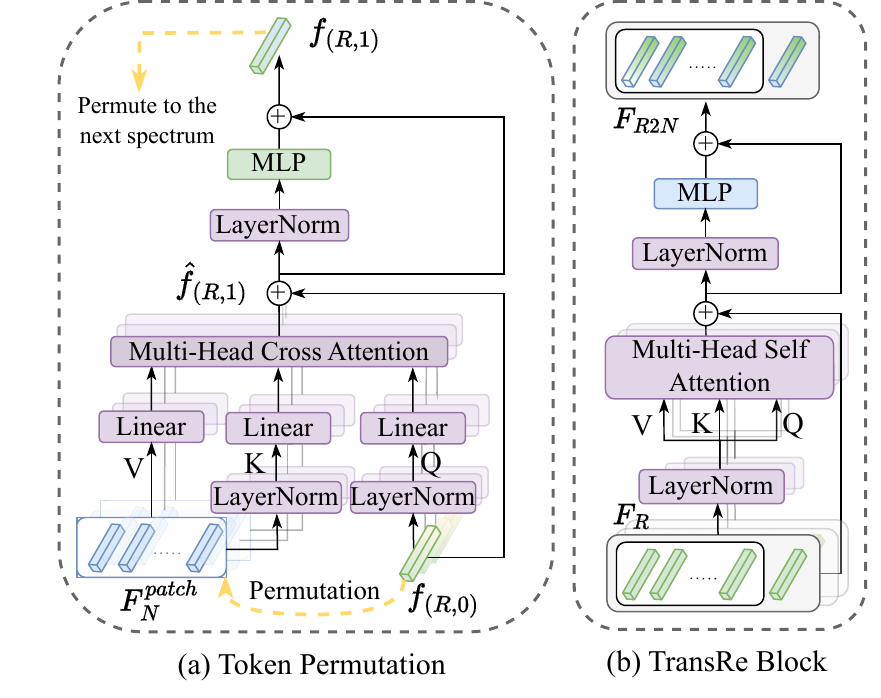}
\caption{Our token permutation and TransRe blocks with the RGB stream. Other streams share a similar structure.}
\label{fig:ca}
\end{figure}
\subsection{Token Permutation Module}
To achieve the spatial feature alignment among different image spectra and the effective aggregation of heterogeneous features, we introduce the Token Permutation Module (TPM) with a cyclic token permutation mechanism, as illustrated at the top right corner of Fig. \ref{fig:Overall}.

Technically, TPM takes the token features $F_{\mathrm{R}}$, $F_{\mathrm{N}}$ and $F_{\mathrm{T}}$ as inputs, and generates the fused feature $f_\mathrm{tp}$ with three consecutive token permutations.
Without loss of generality, we take the RGB stream as a starting example.
As shown in Fig.~\ref{fig:ca} (a), we utilize a Multi-Head Cross-Attention (MHCA)~\cite{dosovitskiy2020image} with $N_h$ heads to achieve the token permutation.
More specifically, the class token ${f}_{\mathrm {(R,0)}} \in R^{D}$ from $F_{\mathrm{R}}$ is passed into a linear transformation to generate a query matrix $Q \in R^{D}$.
The patch tokens $F_{\mathrm {N}}^{patch} \in R^{D \times M}$ from $F_\mathrm {N}$ are passed into two linear transformations to generate a key matrix $K$ and a value matrix $V$, respectively.
Thus, the interaction of $F_{\mathrm{R}}$ and $F_{\mathrm{N}}$ in the $h$-th head is represented as
\begin{equation}
\hat{f}_{\mathrm {(R,1)}}^h = \sigma (\frac{{{Q}^h}{{K}^h}^ \top}{\sqrt{d}}){{V}^h},
\end{equation}
where $\sigma$ is the softmax function and $(\cdot)^\top$ means the matrix transposition.
Here, ${{Q}^h} \in R^{d}$, ${{K}^h}$, ${{V}^h} \in R^{d\times M}$, $d= \frac{D}{N_h}$.
The outputs of $N_h$ heads ($\hat{f}_{\mathrm {(R,1)}}^1,\cdots, \hat{f}_{\mathrm {(R,1)}}^h,\cdots, \hat{f}_{\mathrm {(R,1)}}^{N_h}$) are concatenated to be $\hat{f}_{\mathrm {(R,1)}} \in R^{D}$.
Then, $\hat{f}_{\mathrm {(R,1)}}$ is passed through a Feed-Forward Network (FFN) to generate a new class token ${f}_{\mathrm {(R,1)}}$,
\begin{equation}
{f}_{\mathrm {(R,1)}} = \mathrm{FFN}(\hat{f}_{\mathrm {(R,1)}}) + \hat{f}_{\mathrm {(R,1)}}.
\end{equation}
It serves as the initial spatial alignment of the RGB and NIR image features. Similar operations can be performed for other spectra,
\begin{equation}
    {f}_{\mathrm {(R,1)}}=\mathrm {FFN}(\mathrm {MHCA}\left(\mathrm{LN}({f}_{\mathrm {(R,0)}}), \mathrm{LN}(F_{\mathrm {N}}^{patch})\right)),
\end{equation}
\begin{equation}
    {f}_{\mathrm {(N,1)}}=\mathrm {FFN}(\mathrm { MHCA }\left(\mathrm{LN}({f}_{\mathrm {(N,0)}}), \mathrm{LN}(F_{\mathrm {T}}^{patch})\right)),
\end{equation}
\begin{equation}
    {f}_{\mathrm {(T,1)}}=\mathrm {FFN}(\mathrm { MHCA }\left(\mathrm{LN}({f}_{\mathrm {(T,0)}}), \mathrm{LN}(F_{\mathrm {R}}^{patch})\right)).
\end{equation}
As shown in Fig. \ref{fig:ca} (a) and above equations, we additionally introduce the LayerNorm (LN)~\cite{ba2016layer} to $Q$ and $K$ to ensure the numerical stability.
Thus, the first token permutation can be totally formulated as
\begin{equation}
    {f}_{\mathrm {(R,1)}}, {f}_{\mathrm {(N,1)}}, {f}_{\mathrm {(T,1)}}=\mathrm {TPM^1}(F_{\mathrm{R}},F_{\mathrm{N}},F_{\mathrm{T}}),
\end{equation}
From the above equations, it can be observed that the token permutation enables the global class token from each spectrum to interact with the local patch tokens of the next spectrum, achieving the initial feature fusion and alignment.

Furthermore, the permutated class tokens ${f}_{\mathrm {(R,1)}}$, ${f}_{\mathrm {(N,1)}}$, and ${f}_{\mathrm {(T,1)}}$ are paired with their initial patch tokens to form $ F_{\mathrm{R}\rightarrow\mathrm{N}}$, $F_{\mathrm{N}\rightarrow\mathrm{T}}$, and $F_{\mathrm{T}\rightarrow\mathrm{R}}$, respectively.
The class tokens keep shifting to the next spectrum,
\begin{equation}
{f}_{\mathrm {(R,2)}}, {f}_{\mathrm {(N,2)}}, {f}_{\mathrm {(T,2)}}=\mathrm {TPM^2}(F_{\mathrm{R}\rightarrow\mathrm{N}},F_{\mathrm{N}\rightarrow\mathrm{T}},F_{\mathrm{T}\rightarrow\mathrm{R}}).
\end{equation}
At this stage, each spectrum has already incorporated detail information from other spectra.
Similar to the previous step, the permutated class tokens ${f}_{\mathrm {(R,2)}}$, ${f}_{\mathrm {(N,2)}}$, and ${f}_{\mathrm {(T,2)}}$ are paired with permutated patch tokens to form $F_{\mathrm{RN}\rightarrow\mathrm{T}}$, $F_{\mathrm{NT}\rightarrow\mathrm{R}}$, and $F_{\mathrm{TR}\rightarrow\mathrm{N}}$, respectively.
Finally, the token permutation process ends with each class token interacting with its own patch tokens,
\begin{equation}
{f}_{\mathrm {(R,3)}}, {f}_{\mathrm {(N,3)}}, {f}_{\mathrm {(T,3)}}=\mathrm {TPM^3}(F_{\mathrm{RN}\rightarrow\mathrm{T}},F_{\mathrm{NT}\rightarrow\mathrm{R}}, F_{\mathrm{TR}\rightarrow\mathrm{N}}).
\end{equation}
Through the above token permutation, the information from all other spectra is conveyed to the patch tokens through the class token, enabling robust feature alignment.
Finally, we concatenate the permutated class tokens to obtain the permutated representation $f_{\text{tp}} \in R^{3D}$,
\begin{equation}
f_\mathrm{tp}=\mathrm{Concat}\left({f}_{\mathrm {(R,3)}}, {f}_{\mathrm {(N,3)}}, {f}_{\mathrm {(T,3)}}\right).
\end{equation}
This cyclic token permutation enhances the spatial fusion and implicit alignment of deep features across spectra, improving the ability of inter-spectral dependencies.
\subsection{Complementary Reconstruction Module}
There are significant distribution gaps among different image spectra.
In addition, it often involves the absence of certain image spectra in real world.
Inspired by the image generation~\cite{zhu2017unpaired}, we propose a Complementary Reconstruction Module (CRM) to reduce the distribution gap across different image spectra.
The key is to incorporate dense token-level reconstruction constraints.
%

Without loss of generality, we take the RGB stream as an example and consider the NIR and TIR spectra missing.
To reconstruct the missing tokens, we pass $F_{R}$ through a Transformer-based Reconstruction (TransRe) block (See Fig.~\ref{fig:ca} (b)) and generate the corresponding tokens by
\begin{equation}
F_\mathrm{R2N} = \mathrm {TransRe}(F_\mathrm{R}),
\end{equation}
\begin{equation}
F_\mathrm{R2T} = \mathrm {TransRe}(F_\mathrm{R}),
\end{equation}
where $F_\mathrm{R2N}$, $F_\mathrm{R2T} \in R^{D \times (M+1)}$ are the reconstructed tokens.
The reconstructed tokens $F_\mathrm{R2N}$ and $F_\mathrm{R2T}$ are constrained by the real token features $F_\mathrm{N}$ and $F_\mathrm{T}$ using the Mean Squared Error (MSE) loss:
\begin{equation}
\mathcal{L}_{\mathrm{R2N}} = \frac{1}{M+1}\sum_{i=1}^{M+1}{||F_\mathrm{R2N} - F_\mathrm{N}||_2^2},
\end{equation}
\begin{equation}
    \mathcal{L}_{\mathrm{R2T}} = \frac{1}{M+1}\sum_{i=1}^{M+1}{||F_\mathrm{R2T} - F_\mathrm{T}||_2^2},
\end{equation}
\begin{equation}
    \mathcal{L}_{\mathrm{R}} = \mathcal{L}_{\mathrm{R2N}} + \mathcal{L}_{\mathrm{R2T}}.
\end{equation}
Through the above token-level reconstruction constraints, the distribution gap between RGB and other spectra is reduced.
To improve the reconstruction ability, we introduce similar constraints to all the image spectra and achieve a multi-spectral complementary reconstruction.
The complementary reconstruction loss $\mathcal{L}_{cr}$ can be expressed as the sum of the individual losses for each spectrum:
\begin{equation}
\mathcal{L}_{cr} = \mathcal{L}_{\mathrm{R}} + \mathcal{L}_{\mathrm{N}}+ \mathcal{L}_{\mathrm{T}}.
\end{equation}
By introducing token-level constraints, our CRM effectively reduces the distribution gap among different image spectra.
Moreover, it can generate corresponding tokens of missing spectra, ensuring a unified learning framework even in scenarios where one or more spectra are absent.
\subsection{Dynamic Cooperation between CRM and TPM}
In this work, we further introduce the dynamic cooperation between the CRM and TPM to handle the absence of any image spectrum.
For example, when the RGB image spectrum is missing, the token features $F_\mathrm{N}$ and $F_\mathrm{T}$ will activate their reconstruction blocks to generate the corresponding RGB token features ${F}_\mathrm{N2R}$ and ${F}_\mathrm{T2R}$, respectively.
Then, the reconstructed RGB token features can be represented as
\begin{equation}
\bar{{F}}_\mathrm{R} = \frac{({F}_\mathrm{N2R} + {F}_\mathrm{T2R})}{2}.
\end{equation}
Then, $\bar{F}_\mathrm{R}$, $F_\mathrm{N}$ and $F_\mathrm{T}$ can be fed into the TPM to perform the token permutation as normal.
Hence, our CRM can dynamically cooperate with TPM, ensuring that the missing spectrum can still participate in the permutation process.
\subsection{Objective Function}
As illustrated in Fig. \ref{fig:Overall}, our objective function comprises three components: loss for the ViT backbone, loss for the token permutation and loss for the CRM.
As for the ViT backbone and the token permutation, they are both supervised by the label smoothing cross-entropy loss \cite{szegedy2016rethinking} and triplet loss \cite{hermans2017defense}.
Finally, the total loss in our framework can be defined by
\begin{equation}
\begin{split}
    \mathcal{L}_{total} = \mathcal{L}_{tri}^{ViT} + \mathcal{L}_{ce}^{ViT} + \mathcal{L}_{tri}^{TP} + \mathcal{L}_{ce}^{TP} + \mathcal{L}_{cr}.
\end{split}
\end{equation}
\section{Experiments}
\subsection{Dataset and Evaluation Protocols}
To evaluate the performance, we adopt three multi-spectral object ReID datasets.
RGBNT201~\cite{zheng2021robust} is the first multi-spectral person ReID dataset with RGB, NIR and TIR spectra.
RGBNT100~\cite{li2020multi} is a large-scale multi-spectral vehicle ReID dataset.
MSVR310~\cite{zheng2022multi} is a small-scale multi-spectral vehicle ReID dataset with more complex scenarios.
Following previous works, we adopt the mean Average Precision (mAP) and Cumulative Matching Characteristics (CMC) at Rank-K ($K=1,5,10$) as our evaluation metrics.
\begin{table}[t]
    \centering
    \renewcommand\arraystretch{1.12}
    \setlength\tabcolsep{4pt}
    \resizebox{0.38\textwidth}{!}
	{
    \begin{tabular}{cc|cccc}
    \hline
    \multicolumn{2}{c|}{\multirow{2}{*}{\textbf{Methods}}}     &\multicolumn{4}{c}{\textbf{RGBNT201}}\\ \cline{3-6}
    & & \textbf{mAP} & \textbf{R-1} & \textbf{R-5} & \textbf{R-10} \\ \hline
    \multirow{6}{*}{\textbf{Single}}
    &HACNN & 21.3 & 19.0 & 34.1 & 42.8 \\
    &MUDeep & 23.8 & 19.7 & 33.1 & 44.3 \\
    &OSNet  & 25.4 & 22.3 & 35.1 & 44.7 \\
    &MLFN  & 26.1 & 24.2 & 35.9 & 44.1 \\
    &CAL  & 27.6 & 24.3 & 36.5 & 45.7 \\
    &PCB  & 32.8 & 28.1 & 37.4 & 46.9 \\ \hline
    \multirow{5}{*}{\textbf{Multi}}
    & HAMNet   & 27.7         & 26.3            & 41.5            & 51.7             \\
    & PFNet    & 38.5         & 38.9            & 52.0              & 58.4             \\
    & DENet    & 42.4         & 42.2            & 55.3            & \underline{64.5}             \\
    & IEEE     & \underline{47.5}         & \underline{44.4}            & \underline{57.1}            & 63.6             \\
    & \textbf{TOP-ReID*} & \textbf{72.3}        & \textbf{76.6}           & \textbf{84.7}         & \textbf{89.4}             \\ \hline
    \end{tabular}
    }
    \caption{Performance comparison on RGBNT201.
    The best and second results are in bold and underlined, respectively.
    * signifies Transformer-based approaches, while others are CNN-based ones.}
    \label{tab:multi-spectral person ReID}
\end{table}
\begin{table}[t]
    \centering
    \renewcommand\arraystretch{1.12}
    \setlength\tabcolsep{5pt}
    \resizebox{0.38\textwidth}{!}
	{
    \begin{tabular}{cc|cc|cc}
    \hline
    \multicolumn{2}{c|}{\multirow{2}{*}{\textbf{Methods}}} &  \multicolumn{2}{c|}{\textbf{RGBNT100}} & \multicolumn{2}{c}{\textbf{MSVR310}} \\
    \cline{3-6}
    & & \textbf{mAP} & \textbf{R-1} & \textbf{mAP} & \textbf{R-1} \\
    \hline
    \multirow{9}{*}{\textbf{Single}}
    &DMML& 58.5 & 82.0 & 19.1 & 31.1 \\
    &Circle Loss & 59.4 & 81.7 & 22.7 & 34.2 \\
    &PCB& 57.2 & 83.5 & 23.2 & 42.9 \\
    &MGN & 58.1 & 83.1 & 26.2 & 44.3 \\
    &BoT & 78.0 & 95.1 & 23.5 & 38.4 \\
    &HRCN & 67.1 & 91.8 & 23.4 & 44.2 \\
    &OSNet& 75.0 & 95.6 & 28.7 & 44.8 \\
    &AGW & 73.1 & 92.7 & 28.9 & \underline{46.9} \\
    &TransReID*& 75.6 & 92.9 & 18.4 & 29.6 \\
    \hline
    \multirow{7}{*}{\textbf{Multi}}
    &GAFNet & 74.4 & 93.4 & - & - \\
    &GPFNet & 75.0 & 94.5 & - & - \\
    &PHT* & \underline{79.9} & 92.7 & - & - \\
    &PFNet& 68.1 & 94.1 & 23.5 & 37.4 \\
    &HAMNet & 74.5 & 93.3 & 27.1 & 42.3 \\
    &CCNet & 77.2 & \underline{96.3} & \textbf{36.4} & \textbf{55.2} \\
    & \textbf{TOP-ReID*} & \textbf{81.2} & \textbf{96.4} & \underline{35.9} & 44.6 \\
    \hline
    \end{tabular}
    }
    \caption{Performance on RGBNT100 and MSVR310.}
    \label{tab:multi-spectral vehicle ReID}
\end{table}
\subsection{Implementation Details}
Our model is implemented with the PyTorch toolbox.
We conduct experiments with one NVIDIA A800 GPU.
We use pre-trained Transformers on the ImageNet classification dataset \cite{deng2009imagenet} as our backbones.
All images are resized to 256$\times$128$\times$3 pixels.
When training, random horizontal flipping, cropping and erasing \cite{zhong2020random} are used as data augmentation.
We set the mini-batch size to 128.
Each mini-batch consists of 8 randomly selected object identities, and 16 images are sampled for each identity.
We use the Stochastic Gradient Descent (SGD) optimizer with a momentum coefficient of 0.9 and a weight decay of 0.0001.
Furthermore, the learning rate is initialized as 0.009.
The warmup strategy and cosine decay are used during training.
\subsection{Comparison with State-of-the-Art Methods}
\textbf{Multi-spectral Person ReID.}
In Tab.~\ref{tab:multi-spectral person ReID}, we compare our TOP-ReID with both single-spectral methods and multi-spectral methods on RGBNT201.
The results indicate that single-spectral methods generally achieve lower performance compared with multi-spectral methods.
It demonstrates the effectiveness of utilizing complementary information from different image spectra.
Among the single-spectral methods, PCB achieves the highest performance, attaining the mAP and Rank-1 accuracy of 32.8\% and 28.1\%, respectively.
As for the multi-spectral methods, our TOP-ReID achieves remarkable performance.
Specifically, it achieves a mAP that is 24.8\% higher and a Rank-1 accuracy that surpasses IEEE by 32.2\%.
These performance gains provide strong evidences for our TOP-ReID in tackling the challenges of multi-spectral person ReID.
\begin{table*}[t]
\centering
\renewcommand\arraystretch{1.12}
\setlength\tabcolsep{5pt}
\resizebox{0.8\textwidth}{!}
{
\begin{tabular}{cc|cccccccccccc}
\hline
\multicolumn{2}{c|}{\multirow{2}{*}{\textbf{Methods}}} &  \multicolumn{2}{c}{\textbf{M (RGB)}} & \multicolumn{2}{c}{\textbf{M (NIR)}} & \multicolumn{2}{c}{\textbf{M (TIR)}} & \multicolumn{2}{c}{\textbf{M (RGB+NIR)}} & \multicolumn{2}{c}{\textbf{M (RGB+TIR)}} & \multicolumn{2}{c}{\textbf{M (NIR+TIR)}} \\
\cline{3-14}
& & \textbf{mAP} & \textbf{R-1}  & \textbf{mAP} & \textbf{R-1} & \textbf{mAP} & \textbf{R-1} & \textbf{mAP} & \textbf{R-1} & \textbf{mAP} & \textbf{R-1} & \textbf{mAP} & \textbf{R-1} \\
 \hline
 \multirow{6}{*}{\textbf{Single}} &HACNN  & 12.5 & 11.1 & 20.5 & 19.4 & 16.7 & 13.3 & 9.2 & 6.2 & 6.3 & 2.2 & 14.8 & 12.0 \\
 &MUDeep  & 19.2 & 16.4 & 20.0 & 17.2 & 18.4 & 14.2 & 13.7 & 11.8 & 11.5 & 6.5 & 12.7 & 8.5 \\
 &OSNet  & 19.8 & 17.3 & 21.0 & 19.0 & 18.7 & 14.6 & 12.3 & 10.9 & 9.4 & 5.4 & 13.0 & 10.2 \\
 &MLFN & 20.2 & 18.9 & 21.1 & 19.7 & 17.6 & 11.1 & 13.2 & 12.1 & 8.3 & 3.5 & 13.1 & 9.1 \\
 &CAL  & 21.4 & 22.1 & 24.2 & 23.6 & 18.0 & 12.4 & 18.6 & 20.1 & 10.0 & 5.9 & 17.2 & 13.2 \\
 &PCB  & 23.6 & 24.2 & 24.4 & 25.1 & 19.9 & 14.7 & 20.6 & 23.6 & 11.0 & 6.8 & 18.6 & 14.4 \\
 \hline
 \multirow{3}{*}{\textbf{Multi}} & PFNet  & - & - & 31.9 & 29.8 & 25.5 & 25.8 & - & - & - & - & 26.4 & 23.4 \\
 & DENet  & - & - & \underline{35.4} & \underline{36.8} & \underline{33.0} & \underline{35.4} & - & - & - & - & \underline{32.4} & \underline{29.2} \\
 & \textbf{TOP-ReID} & \textbf{54.4} & \textbf{57.5} & \textbf{64.3} & \textbf{67.6} & \textbf{51.9} & \textbf{54.5} & \textbf{35.3} & \textbf{35.4} & \textbf{26.2} & \textbf{26.0} & \textbf{34.1} & \textbf{31.7} \\
 \hline
\end{tabular}
}
\caption{Experimental results of missing-spectral tasks on RGBNT201. “M (X)" stands for missing the X image spectra.}
\label{tab:missing-spectral person ReID}
\end{table*}
\begin{table}[t]
    \centering
    \renewcommand\arraystretch{1.12}
    \setlength\tabcolsep{4.5pt}
    \resizebox{0.38\textwidth}{!}
	{
    \begin{tabular}{c|cccc|cccc}
        \hline
        &\multicolumn{4}{c|}{\textbf{Modules}}        &\multicolumn{4}{c}{\textbf{RGBNT201}} \\ \cline{2-9}
        & \textbf{BL}& \textbf{AL}& \textbf{TPM}& \textbf{CRM}        & \textbf{mAP} & \textbf{R-1} & \textbf{R-5} & \textbf{R-10} \\
        \hline
        A & \ding{51}& \ding{53}& \ding{53}& \ding{53}& 55.9&	54.9&	70.8&	77.6\\
        B & \ding{53}& \ding{51}& \ding{53}& \ding{53}& 62.9 & 64.5 & 77.4 & 82.7 \\
        C & \ding{53}& \ding{51}& \ding{51}& \ding{53}& 67.8 & 69.4 & 83.3 & 88.8 \\
        D & \ding{53}& \ding{51}& \ding{51}& \ding{51}& 72.3 & 76.6 & 84.7 & 89.4 \\
        \hline
    \end{tabular}
    }
    \caption{Performance comparison with different modules.}
    \label{tab:ablation}
\end{table}

\textbf{Multi-spectral Vehicle ReID.}
As shown in Tab.~\ref{tab:multi-spectral vehicle ReID}, single-spectral methods such as OSNet~\cite{zhou2019omni}, AGW \cite{ye2021deep} and TransReID~\cite{he2021transreid}, stand out for their competitive performance.
For multi-spectral methods, CCNet achieves remarkable results across both datasets.
On the RGBNT100 dataset, our TOP-ReID outperforms CCNet with a 4.0\% higher mAP.
On the small-scale MSVR310 dataset, our TOP-ReID maintains competitive performance, showing its versatility and robustness.

\textbf{Evaluation on Missing-spectral Scenarios.}
As shown in Tab.~\ref{tab:missing-spectral person ReID}, all single-spectral methods suffer from performance degradations when image spectra are missing.
Multi-spectral methods demonstrate better robustness compared with single-spectral methods.
Our proposed TOP-ReID achieves remarkable performance even in the presence of missing spectra.
It consistently outperforms both single-spectral and multi-spectral methods in all missing-spectral scenarios, indicating its effectiveness in handling the spectral incompleteness.
In addition, compared with PFNet and DENet, our TOP-ReID is a more flexible and diverse framework to address any spectra missing.
\subsection{Ablation Studies}
To investigate the effect of different components, we further perform a scope of ablation studies on RGBNT201.

\textbf{Effects of Key Modules.}
Tab.~\ref{tab:ablation} illustrates the performance comparison with different modules.
The Model A is the baseline which utilizes the multi-stream ViT-B/16 backbones.
BL means the triplet loss and cross-entropy loss are added before the concatenation of multi-spectral features, while AL means these losses are employed after the feature concatenation.
It can be observed that the AL setting shows better results.
The main reason is that the fused multi-spectral features is more powerful than the simple feature concatenation.
Furthermore, by integrating our TPM, the Model C yields higher performance with mAP of 67.8\% and Rank-1 of 69.4\%.
By introducing CRM, the final model can achieve the best performance with mAP of 72.3\% and Rank-1 of 76.6\%.
These improvements validate the effectiveness of our key modules in handling complex ReID scenarios.
\begin{table*}[t]
    \centering
    \renewcommand\arraystretch{1.12}
    \setlength\tabcolsep{4.5pt}
    \resizebox{0.8\textwidth}{!}
	{
    \begin{tabular}{c|cccc|cccc|cccc}
    \hline
    \multirow{2}{*}{\textbf{Methods}} & \multicolumn{4}{c|}{\textbf{ViT-B/16}} & \multicolumn{4}{c|}{\textbf{DeiT-S/16}} & \multicolumn{4}{c}{\textbf{T2T-ViT-24}} \\ \cline{2-13}
    & \textbf{mAP} & \textbf{R-1} & \textbf{R-5} & \textbf{R-10} & \textbf{mAP} & \textbf{R-1} & \textbf{R-5} & \textbf{R-10} & \textbf{mAP} & \textbf{R-1} & \textbf{R-5} & \textbf{R-10} \\
    \hline
    RGB & 29.0 & 26.2 & 44.5 & 56.1 & 33.3 & 30.6 & 49.5 & 58.0 & 30.1 & 30.3 & 47.2 & 56.6 \\
    NIR & 18.7 & 14.0 & 31.1 & 44.9 & 22.7 & 21.4 & 39.4 & 47.6 & 15.8 & 15.3 & 27.0 & 36.8 \\
    TIR & 33.4 & 32.3 & 52.4 & 63.3 & 27.1 & 26.3 & 41.3 & 51.4 & 34.0 & 36.2 & 52.0 & 62.0 \\
    NIR-TIR & 45.9 & 43.4 & 59.9 & 69.4 & 40.6 & 40.9 & 54.3 & 61.0 & 40.9 & 40.7 & 56.3 & 64.2 \\
    RGB-NIR & 39.0 & 40.2 & 56.6 & 65.7 & 46.7 & 45.0 & 62.6 & 70.0 & 36.3 & 35.2 & 53.8 & 66.3 \\
    RGB-TIR & 52.6 & 53.8 & 69.0 & 78.2 & 49.3 & 47.8 & 64.1 & 72.8 & 49.9 & 51.7 & 66.7 & 73.8 \\
    RGB-TIR-NIR & 55.9 & 54.9 & 70.8 & 77.6 & 55.1 & 53.3 & 67.3 & 76.2 & 52.2 & 51.3 & 64.1 & 74.3 \\
    Baseline (AL) & 62.9 & 64.5 & 77.4 & 82.7 & 59.9 & 61.1 & 73.9 & 80.9 & 56.2 & 60.4 & 73.0 & 78.6 \\
    Baseline (AL) + TPM & 67.8 & 69.4 & 83.3 & 88.8 & 63.0 & 63.9 & 78.1 & 83.9 & 58.2 & 60.8 & 74.9 & 81.4 \\
    \textbf{Baseline (AL) + CRM + TPM} & \textbf{72.3} & \textbf{76.6} & \textbf{84.7} & \textbf{89.4} & \textbf{69.0} & \textbf{73.6} & \textbf{81.8} & \textbf{84.7} & \textbf{60.0} & \textbf{61.6} & \textbf{76.2} & \textbf{82.3} \\
    \hline
    \end{tabular}
    }
    \caption{Performance comparison of different backbones with different spectra and modules on RGBNT201.}
    \label{tab:different backbones}
\end{table*}

\begin{figure}[t]
\centering
\includegraphics[width=0.42\textwidth]{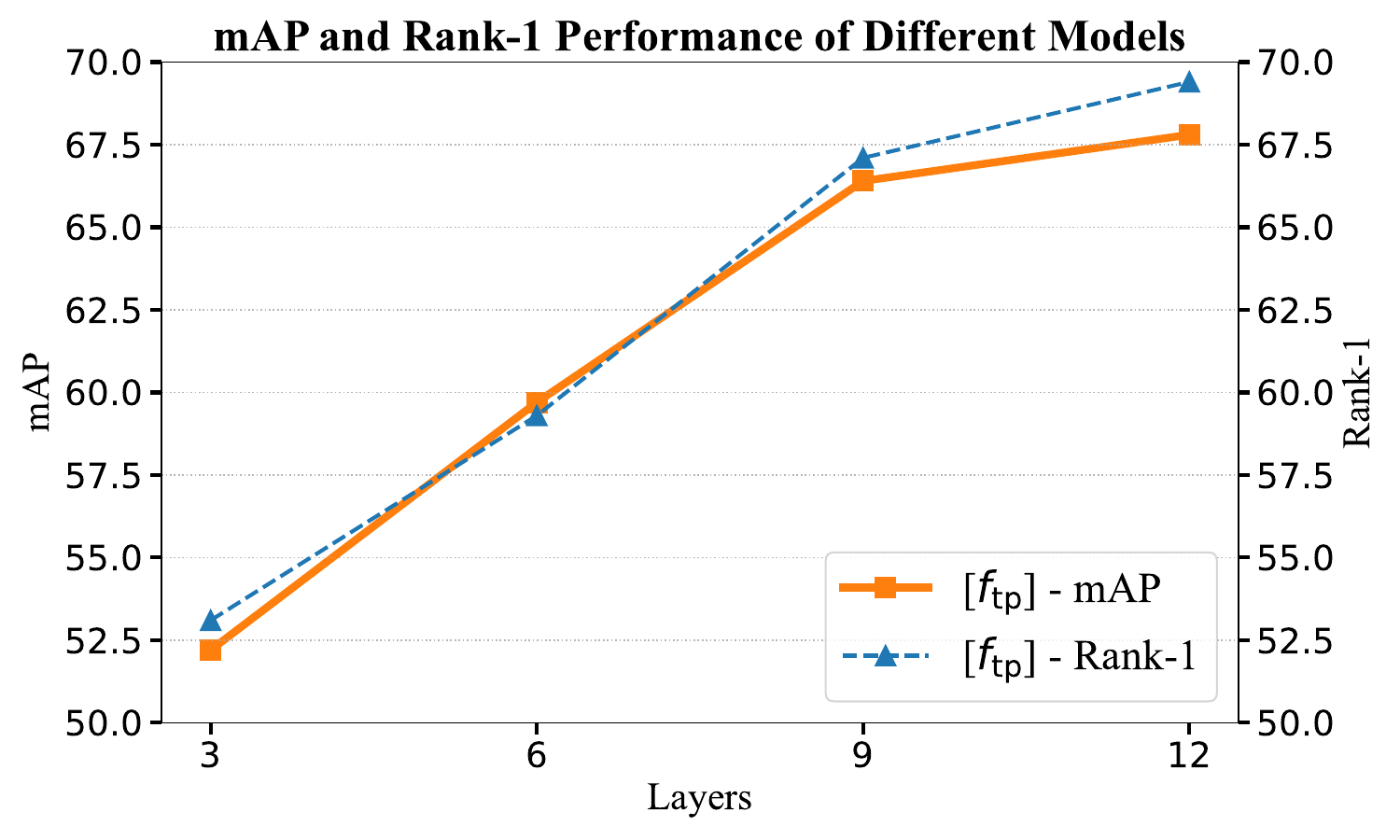}
\caption{Performance of deploying TPM at different layers.}
\label{TPM}
\end{figure}

\textbf{TPM at Different Layers.}
In fact, our TPM is a plug-and-play module.
We explore the effect of TPM at different layers of the ViT backbone.
Fig.~\ref{TPM} shows the performance of TPM at different layers.
We observe that as the plugged depth of TPM increases, the performance greatly improves.
When deployed in the last layer, it achieves the best performance.
This indicates that our TPM is more pronounced in deep layers, capturing more discriminative representations.

\textbf{Effects of TransRe Blocks in CRM.}
\begin{figure}[t]
\centering
\includegraphics[width=0.42\textwidth]{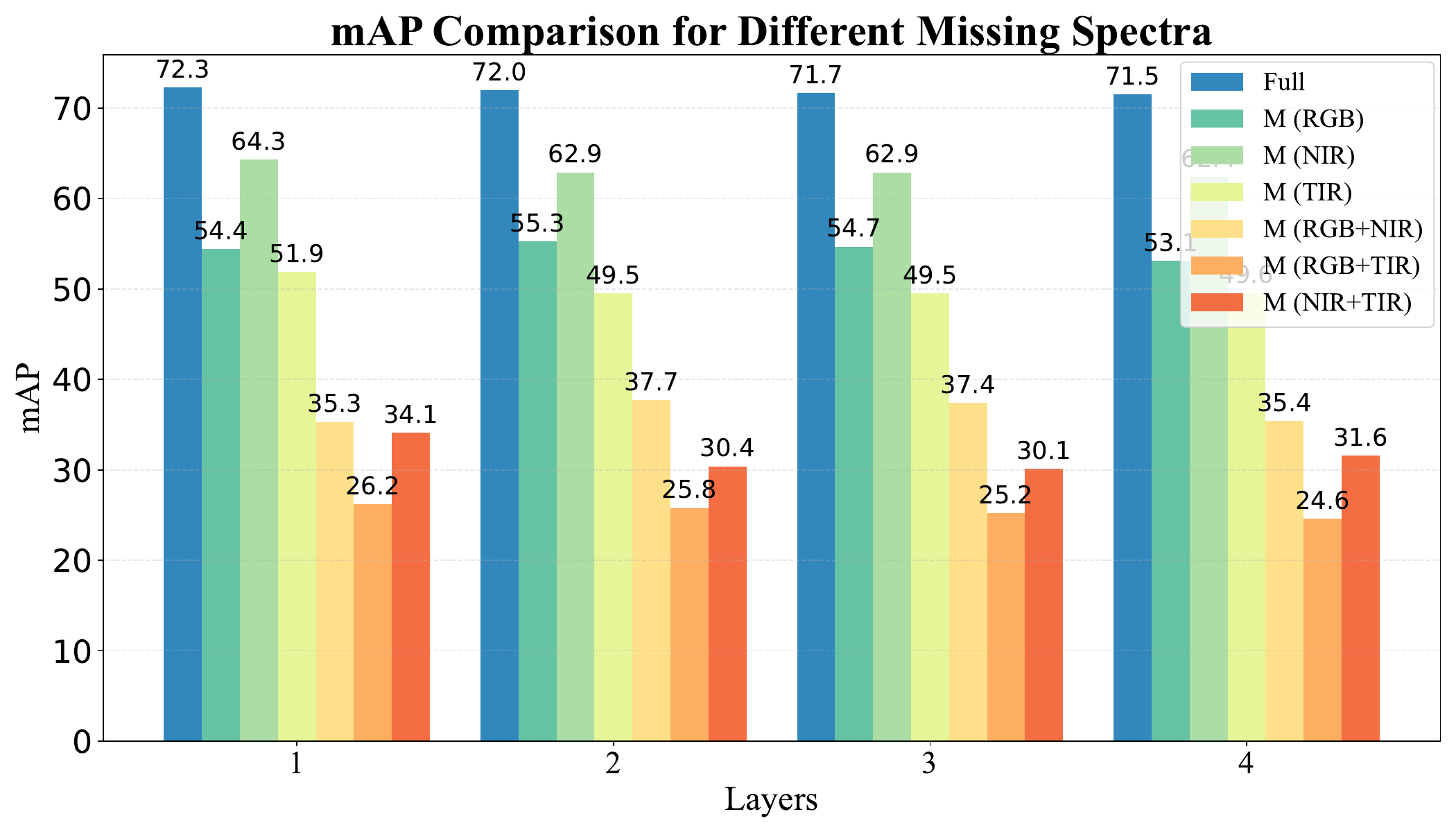}
\caption{Performance comparison with different depths of TransRe blocks in CRM.}
\label{CRM}
\end{figure}
The depth of TransRe blocks may impact the reconstruction ability.
As illustrated in Fig.~\ref{CRM}, the ReID performance is relatively consistent when using different depths of TransRe blocks.
In Fig.~\ref{CRM}, we also provide the comparison results with missing-spectral cases.
It can be observed that the overall performance is acceptable when only using one block.
Thus, we utilize one TransRe block to reduce the computation.

\textbf{Effects of Different Transformer-based Backbones.}
To verify the generalization of our TOP-ReID, we adopt three different Transformer-based backbones, i.e., ViT-B/16, DeiT-S/16 and T2T-ViT-24.
Tab.~\ref{tab:different backbones} illustrates the performance comparison.
As can be observed, the ViT-B/16 delivers the best results.
With more image spectra, different backbones can consistently improve the performance.
Our proposed TPM and CRM can improve the performance with different backbones.
We believe that the performance can be further improved by using more powerful backbones.
\begin{figure}[t]
\centering
\includegraphics[width=0.74\linewidth,height=0.70\linewidth]{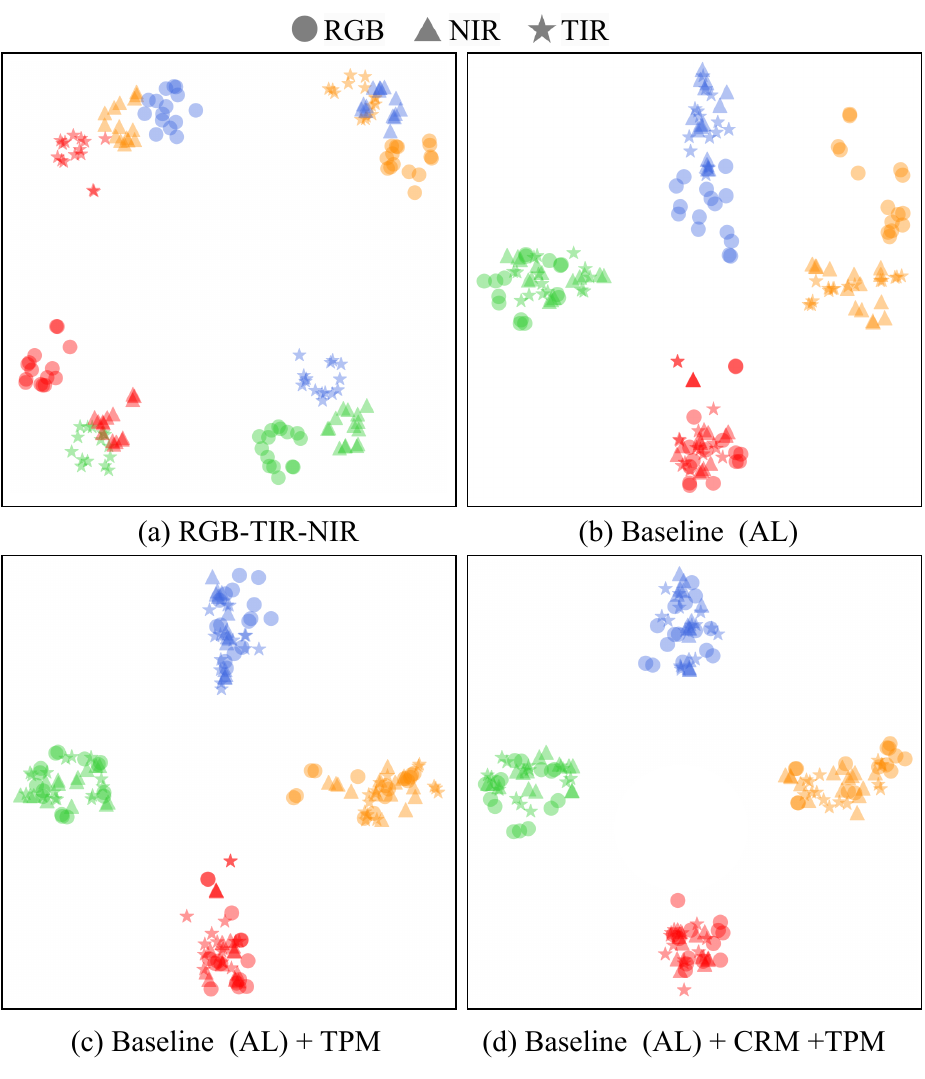}
\caption{Comparison of feature distributions by using t-SNE.
Different colors represent different identities.}
\label{distributions}
\end{figure}
\begin{figure}[t]
\centering
\includegraphics[width=0.40\textwidth]{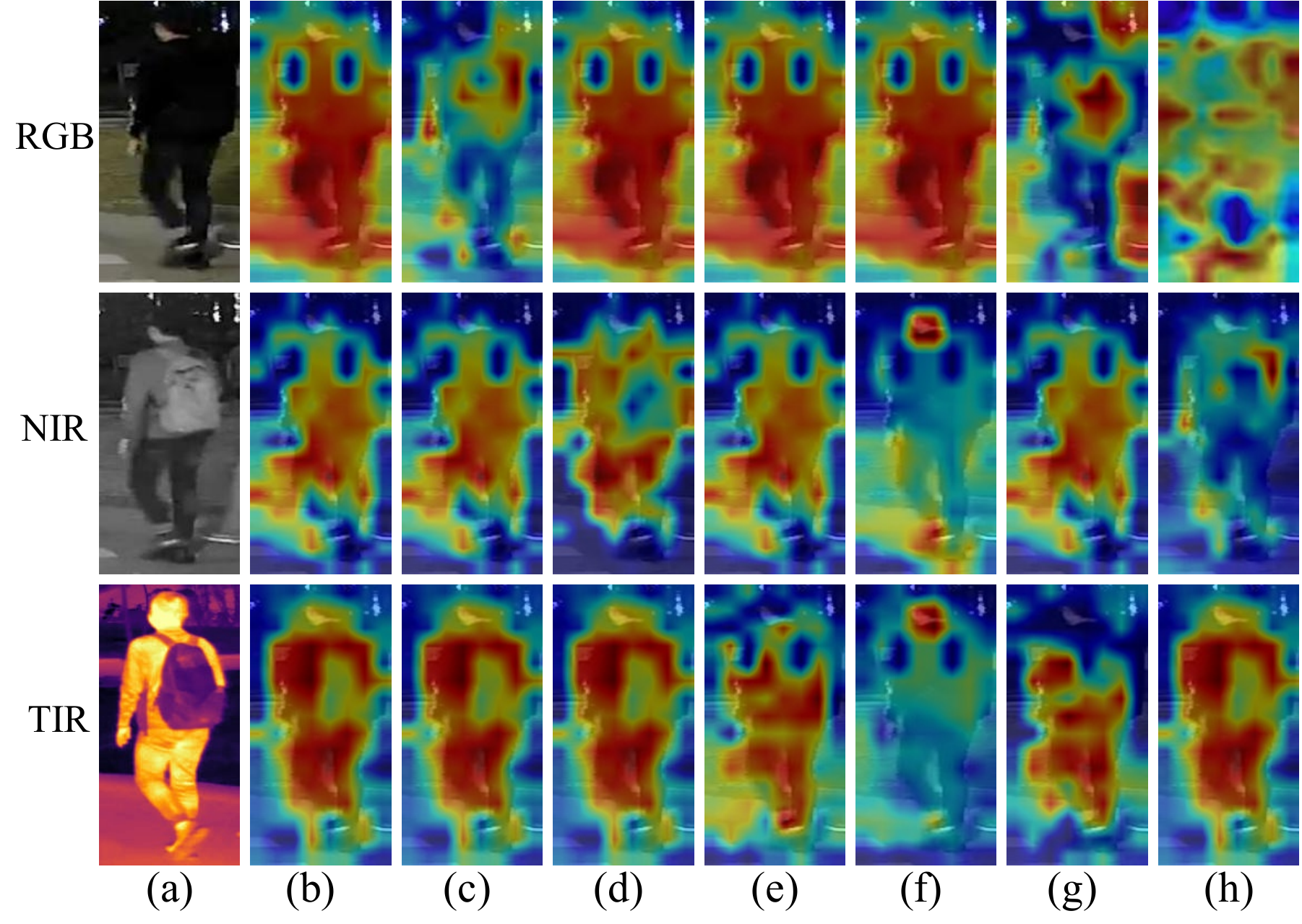}
\caption{Discriminative attention maps. (a) Input images; (b) Full; (c) M (RGB); (d) M (NIR); (e) M (TIR); (f) M (NIR+TIR); (g) M (RGB+TIR); (h) M (RGB+NIR);}
\label{attention_maps}
\end{figure}
\subsection{Visualization Analysis}
To clarify the learning ability, we present visual results on the feature distributions and discriminative attention maps.

\textbf{Multi-spectral Feature Distributions.}
Fig.~\ref{distributions} illustrates the feature distributions of different models by using t-SNE~\cite{van2008visualizing}.
In Fig.~\ref{distributions} (a), it represents the direct concatenation of single-spectral features, where each stream is individually trained.
It can be observed that the AL setting can effectively align the features of different spectra with a better ID consistence.
With our TPM, the features of the same ID across different spectra are more concentrated, and the gaps between different IDs are more distinct.
Furthermore, with CRM, the feature distribution becomes more compact, and the number of outliers for each ID is reduced.
This visualization provides strong evidences for the effectiveness of our proposed methods.

\textbf{Discriminative Attention Maps.}
As shown in Fig.~\ref{attention_maps}, we utilize Grad-CAM~\cite{selvaraju2017grad} to visualize the discriminative attention maps with different image spectra.
Obviously, there are discriminative differences between different image spectra.
Our model is powerful and can highlight discriminative regions when missing image spectra.
%
\section{Conclusion}
In this work, we propose a novel feature learning framework based on token permutations for multi-spectral object ReID.
Our approach incorporates a Token Permutation Module (TPM) for spatial feature alignment and a Complementary Reconstruction Module (CRM) for reducing the distribution gap across different image spectra.
Through the dynamic cooperation between TPM and CRM, it can handle the missing-spectral problem, which is more flexible than previous methods.
Extensive experiments on three benchmarks clearly demonstrate the effectiveness of our methods.

\section{Acknowledgments}
This work was supported in part by the National Key Research and Development Program of China (Grant No. 2018AAA0102001), National Natural Science Foundation of China (Grant No. 62101092), Open Project of Anhui Provincial Key Laboratory of Multimodal Cognitive Computation, Anhui University (Grant No. MMC202102) and Fundamental Research Funds for the Central Universities (No.DUT22QN228).

\bibliography{aaai24}

\begin{thebibliography}{43}
\providecommand{\natexlab}[1]{#1}

\bibitem[{Ba, Kiros, and Hinton(2016)}]{ba2016layer}
Ba, J.~L.; Kiros, J.~R.; and Hinton, G.~E. 2016.
\newblock Layer normalization.
\newblock \emph{arXiv preprint arXiv:1607.06450}.

\bibitem[{Chang, Hospedales, and Xiang(2018)}]{chang2018multi}
Chang, X.; Hospedales, T.~M.; and Xiang, T. 2018.
\newblock Multi-level factorisation net for person re-identification.
\newblock In \emph{CVPR}, 2109--2118.

\bibitem[{Chen et~al.(2019)Chen, Zhang, Lu, and Zhou}]{chen2019deep}
Chen, G.; Zhang, T.; Lu, J.; and Zhou, J. 2019.
\newblock Deep meta metric learning.
\newblock In \emph{ICCV}, 9547--9556.

\bibitem[{Chen et~al.(2022)Chen, Xia, Zhao, Zhou, Niu, Yao, Zhu, and
  Liu}]{chen2022rest}
Chen, Y.; Xia, S.; Zhao, J.; Zhou, Y.; Niu, Q.; Yao, R.; Zhu, D.; and Liu, D.
  2022.
\newblock ResT-ReID: Transformer block-based residual learning for person
  re-identification.
\newblock \emph{PRL}, 157: 90--96.

\bibitem[{Deng et~al.(2009)Deng, Dong, Socher, Li, Li, and
  Fei-Fei}]{deng2009imagenet}
Deng, J.; Dong, W.; Socher, R.; Li, L.-J.; Li, K.; and Fei-Fei, L. 2009.
\newblock Imagenet: A large-scale hierarchical image database.
\newblock In \emph{CVPR}, 248--255.

\bibitem[{Dosovitskiy et~al.(2020)Dosovitskiy, Beyer, Kolesnikov, Weissenborn,
  Zhai, Unterthiner, Dehghani, Minderer, Heigold, Gelly
  et~al.}]{dosovitskiy2020image}
Dosovitskiy, A.; Beyer, L.; Kolesnikov, A.; Weissenborn, D.; Zhai, X.;
  Unterthiner, T.; Dehghani, M.; Minderer, M.; Heigold, G.; Gelly, S.; et~al.
  2020.
\newblock An image is worth 16x16 words: Transformers for image recognition at
  scale.
\newblock \emph{arXiv preprint arXiv:2010.11929}.

\bibitem[{Guo et~al.(2022)Guo, Zhang, Liu, and Wang}]{guo2022generative}
Guo, J.; Zhang, X.; Liu, Z.; and Wang, Y. 2022.
\newblock Generative and attentive fusion for multi-spectral vehicle
  re-identification.
\newblock In \emph{ICSP}, 1565--1572.

\bibitem[{He et~al.(2023)He, Lu, Wang, and Hu}]{he2023graph}
He, Q.; Lu, Z.; Wang, Z.; and Hu, H. 2023.
\newblock Graph-Based Progressive Fusion Network for Multi-Modality Vehicle
  Re-Identification.
\newblock \emph{TITS}, 1--17.

\bibitem[{He et~al.(2021)He, Luo, Wang, Wang, Li, and Jiang}]{he2021transreid}
He, S.; Luo, H.; Wang, P.; Wang, F.; Li, H.; and Jiang, W. 2021.
\newblock Transreid: Transformer-based object re-identification.
\newblock In \emph{ICCV}, 15013--15022.

\bibitem[{Hermans, Beyer, and Leibe(2017)}]{hermans2017defense}
Hermans, A.; Beyer, L.; and Leibe, B. 2017.
\newblock In defense of the triplet loss for person re-identification.
\newblock \emph{arXiv preprint arXiv:1703.07737}.

\bibitem[{Li et~al.(2020{\natexlab{a}})Li, Wei, Hong, and
  Gong}]{li2020infrared}
Li, D.; Wei, X.; Hong, X.; and Gong, Y. 2020{\natexlab{a}}.
\newblock Infrared-visible cross-modal person re-identification with an x
  modality.
\newblock In \emph{AAAI}, volume~34, 4610--4617.

\bibitem[{Li et~al.(2020{\natexlab{b}})Li, Li, Zhu, Zheng, and
  Luo}]{li2020multi}
Li, H.; Li, C.; Zhu, X.; Zheng, A.; and Luo, B. 2020{\natexlab{b}}.
\newblock Multi-spectral vehicle re-identification: A challenge.
\newblock In \emph{AAAI}, volume~34, 11345--11353.

\bibitem[{Li, Zhu, and Gong(2018)}]{li2018harmonious}
Li, W.; Zhu, X.; and Gong, S. 2018.
\newblock Harmonious attention network for person re-identification.
\newblock In \emph{CVPR}, 2285--2294.

\bibitem[{Liu et~al.(2021{\natexlab{a}})Liu, Chai, Tan, Li, and
  Zhou}]{liu2021strong}
Liu, H.; Chai, Y.; Tan, X.; Li, D.; and Zhou, X. 2021{\natexlab{a}}.
\newblock Strong but simple baseline with dual-granularity triplet loss for
  visible-thermal person re-identification.
\newblock \emph{SPL}, 28: 653--657.

\bibitem[{Liu et~al.(2023)Liu, Yu, Zhang, and Lu}]{liu2023deeply}
Liu, X.; Yu, C.; Zhang, P.; and Lu, H. 2023.
\newblock Deeply Coupled Convolution--Transformer With Spatial--Temporal
  Complementary Learning for Video-Based Person Re-Identification.
\newblock \emph{TNNLS}.

\bibitem[{Liu et~al.(2021{\natexlab{b}})Liu, Zhang, Yu, Lu, and
  Yang}]{Liu_2021_CVPR}
Liu, X.; Zhang, P.; Yu, C.; Lu, H.; and Yang, X. 2021{\natexlab{b}}.
\newblock Watching You: Global-Guided Reciprocal Learning for Video-Based
  Person Re-Identification.
\newblock In \emph{Proceedings of the IEEE/CVF Conference on Computer Vision
  and Pattern Recognition (CVPR)}, 13334--13343.

\bibitem[{Luo et~al.(2019)Luo, Gu, Liao, Lai, and Jiang}]{luo2019bag}
Luo, H.; Gu, Y.; Liao, X.; Lai, S.; and Jiang, W. 2019.
\newblock Bag of tricks and a strong baseline for deep person
  re-identification.
\newblock In \emph{CVPRW}, 1487--1495.

\bibitem[{Pan et~al.(2023)Pan, Huang, Liang, Hong, and
  Zhu}]{pan2023progressively}
Pan, W.; Huang, L.; Liang, J.; Hong, L.; and Zhu, J. 2023.
\newblock Progressively Hybrid Transformer for Multi-Modal Vehicle
  Re-Identification.
\newblock \emph{Sensors}, 23(9): 4206.

\bibitem[{Pan et~al.(2022)Pan, Wu, Zhu, Zeng, and Zhu}]{pan2022h}
Pan, W.; Wu, H.; Zhu, J.; Zeng, H.; and Zhu, X. 2022.
\newblock H-vit: Hybrid vision transformer for multi-modal vehicle
  re-identification.
\newblock In \emph{CICAI}, 255--267. Springer.

\bibitem[{Qian et~al.(2017)Qian, Fu, Jiang, Xiang, and Xue}]{qian2017multi}
Qian, X.; Fu, Y.; Jiang, Y.-G.; Xiang, T.; and Xue, X. 2017.
\newblock Multi-scale deep learning architectures for person re-identification.
\newblock In \emph{ICCV}, 5399--5408.

\bibitem[{Qian et~al.(2019)Qian, Fu, Xiang, Jiang, and Xue}]{qian2019leader}
Qian, X.; Fu, Y.; Xiang, T.; Jiang, Y.-G.; and Xue, X. 2019.
\newblock Leader-based multi-scale attention deep architecture for person
  re-identification.
\newblock \emph{TPAMI}, 42(2): 371--385.

\bibitem[{Rao et~al.(2021)Rao, Chen, Lu, and Zhou}]{rao2021counterfactual}
Rao, Y.; Chen, G.; Lu, J.; and Zhou, J. 2021.
\newblock Counterfactual attention learning for fine-grained visual
  categorization and re-identification.
\newblock In \emph{ICCV}, 1025--1034.

\bibitem[{Selvaraju et~al.(2017)Selvaraju, Cogswell, Das, Vedantam, Parikh, and
  Batra}]{selvaraju2017grad}
Selvaraju, R.~R.; Cogswell, M.; Das, A.; Vedantam, R.; Parikh, D.; and Batra,
  D. 2017.
\newblock Grad-cam: Visual explanations from deep networks via gradient-based
  localization.
\newblock In \emph{ICCV}, 618--626.

\bibitem[{Sun et~al.(2020)Sun, Cheng, Zhang, Zhang, Zheng, Wang, and
  Wei}]{sun2020circle}
Sun, Y.; Cheng, C.; Zhang, Y.; Zhang, C.; Zheng, L.; Wang, Z.; and Wei, Y.
  2020.
\newblock Circle loss: A unified perspective of pair similarity optimization.
\newblock In \emph{CVPR}, 6398--6407.

\bibitem[{Sun et~al.(2018)Sun, Zheng, Yang, Tian, and Wang}]{sun2018beyond}
Sun, Y.; Zheng, L.; Yang, Y.; Tian, Q.; and Wang, S. 2018.
\newblock Beyond part models: Person retrieval with refined part pooling (and a
  strong convolutional baseline).
\newblock In \emph{ECCV}, 480--496.

\bibitem[{Szegedy et~al.(2016)Szegedy, Vanhoucke, Ioffe, Shlens, and
  Wojna}]{szegedy2016rethinking}
Szegedy, C.; Vanhoucke, V.; Ioffe, S.; Shlens, J.; and Wojna, Z. 2016.
\newblock Rethinking the inception architecture for computer vision.
\newblock In \emph{CVPR}, 2818--2826.

\bibitem[{Touvron et~al.(2021)Touvron, Cord, Douze, Massa, Sablayrolles, and
  J{\'e}gou}]{touvron2021training}
Touvron, H.; Cord, M.; Douze, M.; Massa, F.; Sablayrolles, A.; and J{\'e}gou,
  H. 2021.
\newblock Training data-efficient image transformers \& distillation through
  attention.
\newblock In \emph{ICML}, 10347--10357.

\bibitem[{Van~der Maaten and Hinton(2008)}]{van2008visualizing}
Van~der Maaten, L.; and Hinton, G. 2008.
\newblock Visualizing data using t-SNE.
\newblock \emph{JMLR}, 9(11).

\bibitem[{Wang et~al.(2018)Wang, Yuan, Chen, Li, and Zhou}]{wang2018learning}
Wang, G.; Yuan, Y.; Chen, X.; Li, J.; and Zhou, X. 2018.
\newblock Learning discriminative features with multiple granularities for
  person re-identification.
\newblock In \emph{ACM MM}, 274--282.

\bibitem[{Wang et~al.(2022{\natexlab{a}})Wang, Shen, Liu, Gao, and
  Gavves}]{wang2022nformer}
Wang, H.; Shen, J.; Liu, Y.; Gao, Y.; and Gavves, E. 2022{\natexlab{a}}.
\newblock Nformer: Robust person re-identification with neighbor transformer.
\newblock In \emph{CVPR}, 7297--7307.

\bibitem[{Wang et~al.(2022{\natexlab{b}})Wang, Li, Zheng, He, and
  Tang}]{wang2022interact}
Wang, Z.; Li, C.; Zheng, A.; He, R.; and Tang, J. 2022{\natexlab{b}}.
\newblock Interact, embed, and enlarge: Boosting modality-specific
  representations for multi-modal person re-identification.
\newblock In \emph{AAAI}, volume~36, 2633--2641.

\bibitem[{Ye et~al.(2021)Ye, Shen, Lin, Xiang, Shao, and Hoi}]{ye2021deep}
Ye, M.; Shen, J.; Lin, G.; Xiang, T.; Shao, L.; and Hoi, S.~C. 2021.
\newblock Deep learning for person re-identification: A survey and outlook.
\newblock \emph{TPAMI}, 44(6): 2872--2893.

\bibitem[{Yuan et~al.(2021)Yuan, Chen, Wang, Yu, Shi, Jiang, Tay, Feng, and
  Yan}]{yuan2021tokens}
Yuan, L.; Chen, Y.; Wang, T.; Yu, W.; Shi, Y.; Jiang, Z.-H.; Tay, F.~E.; Feng,
  J.; and Yan, S. 2021.
\newblock Tokens-to-token vit: Training vision transformers from scratch on
  imagenet.
\newblock In \emph{ICCV}, 558--567.

\bibitem[{Zhang et~al.(2021)Zhang, Zhang, Qi, and Lu}]{zhang2021hat}
Zhang, G.; Zhang, P.; Qi, J.; and Lu, H. 2021.
\newblock Hat: Hierarchical aggregation transformers for person
  re-identification.
\newblock In \emph{ACM MM}, 516--525.

\bibitem[{Zhang and Wang(2023)}]{zhang2023diverse}
Zhang, Y.; and Wang, H. 2023.
\newblock Diverse Embedding Expansion Network and Low-Light Cross-Modality
  Benchmark for Visible-Infrared Person Re-identification.
\newblock In \emph{CVPR}, 2153--2162.

\bibitem[{Zhao et~al.(2021)Zhao, Zhao, Li, Yan, and
  Tian}]{zhao2021heterogeneous}
Zhao, J.; Zhao, Y.; Li, J.; Yan, K.; and Tian, Y. 2021.
\newblock Heterogeneous relational complement for vehicle re-identification.
\newblock In \emph{ICCV}, 205--214.

\bibitem[{Zheng et~al.(2023)Zheng, He, Wang, Li, and Tang}]{zheng2023dynamic}
Zheng, A.; He, Z.; Wang, Z.; Li, C.; and Tang, J. 2023.
\newblock Dynamic Enhancement Network for Partial Multi-modality Person
  Re-identification.
\newblock \emph{arXiv preprint arXiv:2305.15762}.

\bibitem[{Zheng et~al.(2021)Zheng, Wang, Chen, Li, and Tang}]{zheng2021robust}
Zheng, A.; Wang, Z.; Chen, Z.; Li, C.; and Tang, J. 2021.
\newblock Robust multi-modality person re-identification.
\newblock In \emph{AAAI}, volume~35, 3529--3537.

\bibitem[{Zheng et~al.(2022)Zheng, Zhu, Ma, Li, Tang, and Ma}]{zheng2022multi}
Zheng, A.; Zhu, X.; Ma, Z.; Li, C.; Tang, J.; and Ma, J. 2022.
\newblock Multi-spectral vehicle re-identification with cross-directional
  consistency network and a high-quality benchmark.
\newblock \emph{arXiv preprint arXiv:2208.00632}.

\bibitem[{Zhong et~al.(2020)Zhong, Zheng, Kang, Li, and Yang}]{zhong2020random}
Zhong, Z.; Zheng, L.; Kang, G.; Li, S.; and Yang, Y. 2020.
\newblock Random erasing data augmentation.
\newblock In \emph{AAAI}, volume~34, 13001--13008.

\bibitem[{Zhou et~al.(2019)Zhou, Yang, Cavallaro, and Xiang}]{zhou2019omni}
Zhou, K.; Yang, Y.; Cavallaro, A.; and Xiang, T. 2019.
\newblock Omni-scale feature learning for person re-identification.
\newblock In \emph{ICCV}, 3702--3712.

\bibitem[{Zhu et~al.(2017)Zhu, Park, Isola, and Efros}]{zhu2017unpaired}
Zhu, J.-Y.; Park, T.; Isola, P.; and Efros, A.~A. 2017.
\newblock Unpaired image-to-image translation using cycle-consistent
  adversarial networks.
\newblock In \emph{ICCV}, 2223--2232.

\bibitem[{Zhu et~al.(2021)Zhu, Guo, Zhang, Wang, Huang, Qiao, Liu, Wang, and
  Tang}]{zhu2021aaformer}
Zhu, K.; Guo, H.; Zhang, S.; Wang, Y.; Huang, G.; Qiao, H.; Liu, J.; Wang, J.;
  and Tang, M. 2021.
\newblock Aaformer: Auto-aligned transformer for person re-identification.
\newblock \emph{arXiv preprint arXiv:2104.00921}.

\end{thebibliography}
\newpage
\section{A Introduction for Supplementary Material}
In the supplementary material, we provide additional evidence to validate the effectiveness of TOP-ReID.
We perform extra ablation experiments across various datasets and present more visualization results.
The experimental section includes an introduction to comparative methods, validation of different reconstruction methods of CRM, validation of the TPM's effectiveness with three permutations and other details, further examination of key modules within person and vehicle ReID, and comparative experiments in situations where vehicle spectra are absent.
The visualization section includes three parts: the visualization of token alignments, the presentation of Grad-CAM attention maps for both person and vehicle ReID, as well as the showcasing of reconstructed tokens within the CRM module.
\section{B Experiments}
\subsection{B.1 Experimental Details}
In the main text, we primarily conduct comparisons involving multi-spectral and missing-spectral scenarios on the RGBNT201 person ReID dataset, and multi-spectral comparisons on the RGBNT100 vehicle ReID dataset.

\textbf{The single-spectral methods we mainly considered include:}
HACNN \cite{li2018harmonious} proposes a novel approach for jointly learning multi-scale attention selection and feature representation.
MUDeep \cite{qian2019leader} introduces a multi-scale deep learning layer and a leadership-based attention mechanism.
OSNet \cite{zhou2019omni} employs a unified aggregation gate for dynamically fusing omni-scale features.
MLFN \cite{chang2018multi} utilizes a multi-layered structure to extract optimal features at different convolutional depths, which are then integrated for person ReID.
CAL \cite{rao2021counterfactual} introduces a counterfactual attention learning method based on causal reasoning, which learns more effective attention mechanisms.
PCB \cite{sun2018beyond} and MGN \cite{wang2018learning} adapt a stripe-based image division strategy to obtain multi-grained representations.
DMML \cite{chen2019deep} presents a meta perspective on metric learning, demonstrating the consistency of softmax and triplet losses in the meta space.
Circle Loss \cite{sun2020circle} introduces a novel approach to re-weight similarity scores, emphasizing less-optimized similarities for more flexible optimization.
Luo et al. \cite{luo2019bag} utilize a deep residual network and introduce the BNNeck technique.
HRCN \cite{zhao2021heterogeneous} proposes the Cross-camera Generalization Measure (CGM) to enhance evaluations.
AGW \cite{ye2021deep} incorporates non-local attention mechanisms for fine-grained feature extraction.
TransReID \cite{he2021transreid} introduces the first purely Transformer-based method, fully exploiting relationships between image patches.

\textbf{The multi-spectral methods we mainly considered include:}
HAMNet \cite{li2020multi}, designed to effectively fuse various spectral features.
PFNet \cite{zheng2021robust}, which significantly advances multi-spectral person ReID by learning robust RGB-NIR-TIR features.
DENet \cite{zheng2023dynamic}, proposing pixel-level constraints to address the spectral-missing problem.
IEEE \cite{wang2022interact}, a set of three innovative learning strategies aimed at enhancing modality-specific representations.
GAFNet \cite{guo2022generative}, which seamlessly fuses data from multiple sources.
GPFNet, introducing an adaptive fusion approach for multi-spectral features.
HViT \cite{pan2022h} balances modal-specific and modal-shared information.
Furthermore, they employ random hybrid augmentation and a feature hybrid mechanism to enhance performance with PHT \cite{pan2023progressively}.
CCNet, which concurrently addresses discrepancies from both spectra and sample aspects.
These comparisons serve to validate the effectiveness of the proposed method TOP-ReID across various scenarios and datasets.
\subsection{B.2 More Ablation Experiments}
\textbf{Effects of Different Reconstruction Methods.}
Tab. \ref{tab:reconsrturct} shows more comparison results with other reconstruction losses (MAE, RMSE).
The results clearly show that our CRM always surpasses the pixel-level reconstruction method (DENet).
The reason to bridge the distribution gap is that the reconstruction constraints of our CRM promote mutual supervision between different modalities, facilitating the convergence of their feature distributions.
\begin{table}[h]
    \centering
    \renewcommand\arraystretch{1.12}
    \setlength\tabcolsep{4.5 pt}
    \begin{tabular}{c|c|cccc}
    \hline
    \multirow{2}{*}{\textbf{Level}} &\multirow{2}{*}{\textbf{Methods}} & \multicolumn{4}{c}{\textbf{RGBNT201}} \\ \cline{3-6}
    & &\textbf{Full} &\textbf{M(N)} &\textbf{M(T)} & \textbf{M(N+T)} \\
    \hline
    Pixel & DENet &42.4  & 35.4 & 33.0 & 32.4 \\
    Token & CRM (MSE) &\textbf{72.3} & \textbf{64.3} & \underline{51.9} & \underline{34.1}  \\
    Token & CRM (MAE) &68.9 & 59.2 & 51.3 & 32.6  \\
    Token & CRM (RMSE) &\underline{69.8}  & \underline{61.7} & \textbf{54.3} & \textbf{35.8}  \\
    \hline
    \end{tabular}
    \caption{mAP Comparison with different reconstructions.}
    \label{tab:reconsrturct}
\end{table}

\textbf{Evaluation on Three Permutations in TPM.}
As shown in Tab. \ref{tab:permutation}, a sequence of three consecutive permutations showcases how the model progressively generates more decisive representations.
The initial permutation, denoted as the first step, merges each spectrum with its adjacent spectrum to initiate fusion.
Following this, the second permutation step facilitates the incorporation of detailed information from all spectra into one another.
Ultimately, the third permutation step brings the information back into distinct branches, finalizing the fusion process.
This experiment underscores the importance of executing three permutations.
This fusion framework, we believe, not only suits multi-spectral data but can also be effectively applied for efficient information fusion when dealing with other modalities, such as text, sketch, audio, and video.
\begin{table}[h]
    \centering
    \renewcommand\arraystretch{1.24}
    \setlength\tabcolsep{3pt}
    \begin{tabular}{c|ccc|cccc}
        \hline
        &\multicolumn{3}{c|}{\textbf{Modules}}        &\multicolumn{4}{c}{\textbf{RGBNT201}} \\ \cline{2-8}
        & \textbf{TPM$^\textbf{1}$}& \textbf{TPM$^\textbf{2}$}& \textbf{TPM$^\textbf{3}$}       & \textbf{mAP} & \textbf{R-1} & \textbf{R-5} & \textbf{R-10} \\
        \hline
        A & \ding{51}& \ding{53}& \ding{53}& 65.1 &	67.2 & 80.4 & 84.6\\
        B & \ding{51}& \ding{51}& \ding{53}& 66.5 & 69.3 & 81.1 & 86.6 \\
        C & \ding{51}& \ding{51}& \ding{51}& 67.8 & 69.4 & 83.3 & 88.8 \\\hline
    \end{tabular}
    \caption{Performance comparison with consecutive permutations of TPM.}
    \label{tab:permutation}
\end{table}

\textbf{The differences and effects of ${f}_{\mathrm {(R,3)}}, {f}_{\mathrm {(N,3)}}$ and ${f}_{\mathrm {(T,3)}}$.}
(1) Differences: ${f}_{\mathrm {(R,3)}}, {f}_{\mathrm {(N,3)}}$ and ${f}_{\mathrm {(T,3)}}$ are the class tokens representing the RGB, NIR and TIR features after cross-modal fusion, respectively.
They are different. Each of them contains modality-aware information after cross-modal fusion and intra-modal interaction.
(2) Effects: They do not contribute equally to the final result.
As shown in Tab. \ref{tab:diffcls}, ${f}_{\mathrm {(T,3)}}$ contributes more due to the richer detail retention of TIR images.
While ${f}_{\mathrm {(N,3)}}$ contributes less due to the blurriness of NIR images.
\begin{table}[h]
    \centering
    \renewcommand\arraystretch{1.2}
    \setlength\tabcolsep{4.5pt}
    \begin{tabular}{c|cccc}
    \hline
    \multirow{2}{*}{\textbf{Tokens}} & \multicolumn{4}{c}{\textbf{RGBNT201}} \\ \cline{2-5}
    & \textbf{mAP} & \textbf{R-1} & \textbf{R-5} & \textbf{R-10} \\
    \hline
    ${f}_{\mathrm {(R,3)}}$  & 34.1 & 32.2 & 46.1 & 56.1  \\
    ${f}_{\mathrm {(N,3)}}$ & 26.3 & 26.7 & 42.2 & 51.9 \\
    ${f}_{\mathrm {(T,3)}}$ & 35.3 & 35.9 & 53.1 & 60.3 \\
    \hline
    \end{tabular}
    \caption{Effects of ${f}_{\mathrm {(R,3)}}, {f}_{\mathrm {(N,3)}}$ and ${f}_{\mathrm {(T,3)}}$.}
    \label{tab:diffcls}
\end{table}

\begin{figure}[t]
    \centering
    \includegraphics[width=0.45\textwidth]{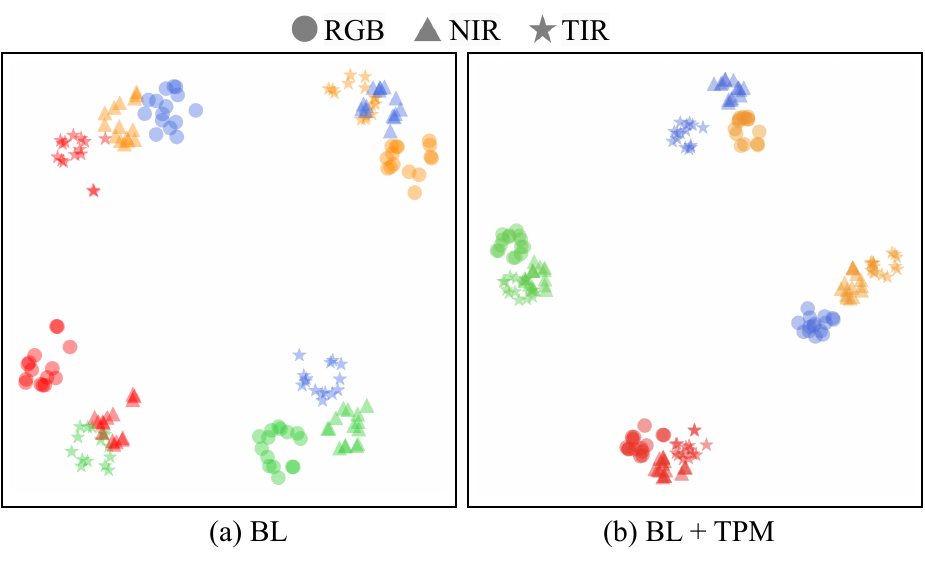}
    \caption{Comparison of feature distributions by using t-SNE.
    Different colors represent different identities.}
    \label{appendix:tsne}
    \end{figure}

\textbf{Evaluation of Key Modules on RGBNT201.}
We evaluate the TPM and CRM with BL, which supervises the features from three ViT-B/16 backbones before the concatenation of features.
Even without utilizing the ID constraint from AL in the main text, we implicitly achieve the alignment of IDs through TPM, as illustrated in Fig. \ref{appendix:tsne}.
The feature distributions for the three spectra corresponding to the same ID become more closely aligned. While some features may not receive strong constraints, it still indicates the potential ID-constraining capability of TPM.

Building upon BL, the introduction of TPM yields a noteworthy 3.5\% increase in mAP and a 6.1\% improvement in rank-1 accuracy.
Furthermore, incorporating CRM, with its dense token constraints, results in a mAP improvement of 5.2\%.
These results demonstrate the effectiveness of TPM and CRM in person ReID.
\begin{table}[t]
    \centering
    \renewcommand\arraystretch{1.12}
    \setlength\tabcolsep{3.6pt}
    \begin{tabular}{c|cccc}
    \hline
    \multirow{2}{*}{\textbf{Methods}} & \multicolumn{4}{c}{\textbf{RGBNT100}} \\ \cline{2-5}
    & \textbf{mAP} & \textbf{R-1} & \textbf{R-5} & \textbf{R-10} \\
    \hline
    RGB & 50.1 & 71.5 & 74.8 & 76.7 \\
    NIR & 43.4 & 61.3 & 64.5 & 66.4 \\
    TIR & 40.9 & 72.1 & 75.0 & 77.3 \\
    NIR-TIR & 65.0 & 88.3 & 89.8 & 91.1 \\
    RGB-NIR & 60.4 & 76.8 & 79.1 & 80.9 \\
    RGB-TIR & 68.1 & 88.7 & 90.3 & 91.2 \\
    Baseline (AL) & 71.6 & 89.4 & 91.0 & 91.8 \\
    Baseline (BL) & 75.2 & 93.9 & 95.2 & 96.2 \\
    Baseline (AL) + TPM & 72.4 & 89.9 & 91.3 & 92.1 \\
    Baseline (BL) + TPM & 78.2 & 95.9 & 97.5 & 98.1 \\
    Baseline (AL) + CRM + TPM & 73.7 & 92.2 & 93.6 & 94.1 \\
    \textbf{Baseline (BL) + CRM + TPM} & \textbf{81.2} & \textbf{96.4} & \textbf{96.9} & \textbf{97.2} \\
    \hline
    \end{tabular}
    \caption{Performance comparison of different modules.}
    \label{tab:ViT-B/16 results on RGBNT100}
\end{table}
\begin{table}[t]
    \centering
    \renewcommand\arraystretch{1.24}
    \setlength\tabcolsep{3pt}
    \begin{tabular}{c|ccc|cccc}
        \hline
        &\multicolumn{3}{c|}{\textbf{Modules}}        &\multicolumn{4}{c}{\textbf{RGBNT201}} \\ \cline{2-8}
        & \textbf{BL}& \textbf{TPM}& \textbf{CRM}       & \textbf{mAP} & \textbf{R-1} & \textbf{R-5} & \textbf{R-10} \\
        \hline
        A & \ding{51}& \ding{53}& \ding{53}& 55.9&	54.9&	70.8&	77.6\\
        B & \ding{51}& \ding{51}& \ding{53}& 59.4 & 61.0 & 76.4 & 81.7 \\
        C & \ding{51}& \ding{51}& \ding{51}& 64.6 & 64.6 & 77.4 & 82.4 \\\hline
    \end{tabular}
    \caption{Performance comparison with BL on RGBNT201.}
    \label{tab:RGBNT201}

\end{table}

\textbf{Evaluation of Key Modules on RGBNT100.}
In the main text, we specifically conduct ablation experiments focusing on person ReID.
However, for a comprehensive evaluation of our proposed method, we also carry out corresponding ablation experiments for vehicle ReID.
In these experiments, we introduce two variants: Baseline (BL) and Baseline (AL).
BL applies the loss before concatenating features from the three branches, while AL applies the loss after concatenation.
Interestingly, in the context of vehicle ReID, introducing AL does not significantly enhance the model's representational capability, unlike in person ReID.
Although AL can improve ID correspondence consistency, separate branch training in AL is less effective compared to BL in vehicle ReID.
In fact, TPM's features implicitly achieve ID correspondence consistency.
Consequently, we uniformly adopt BL to enhance the discriminative representation of each branch in vehicle ReID.

For both AL and BL, the incorporation of TPM and CRM consistently yields substantial improvements.
Notably, when each branch demonstrates robust representational capability, the TPM and CRM impart even more pronounced enhancements.
This compellingly substantiates the effectiveness of TOP-ReID for vehicle ReID.

\begin{figure}[h]
    \centering
    \includegraphics[width=0.48\textwidth]{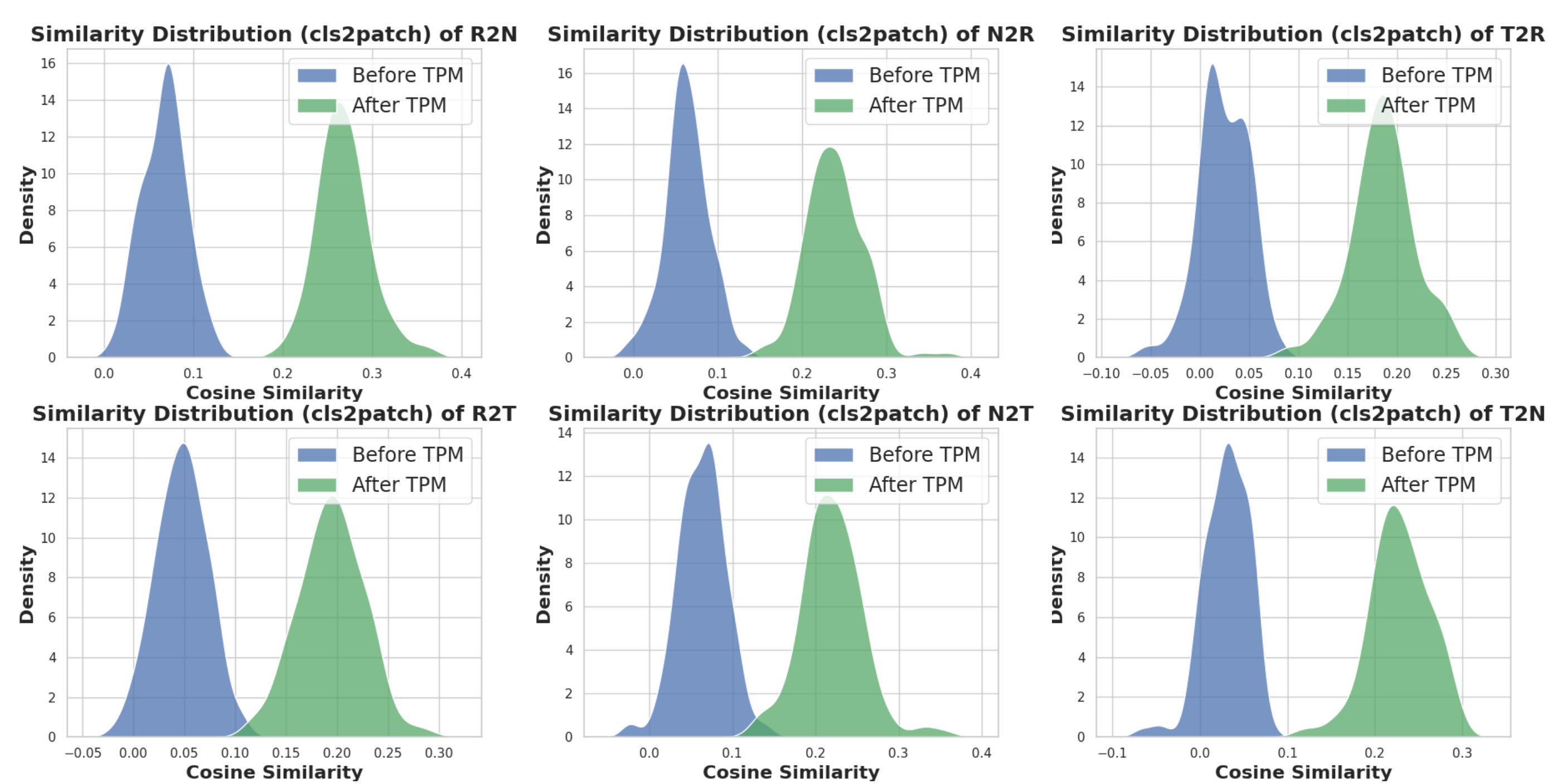}
    \caption{Alignment evidences in TPM. Zoom to see better.}
    \label{fig:align}
    \end{figure}
\textbf{Evaluation on Missing-spectral Scenarios.}
We also assess the effectiveness of our model using the RGBNT100 dataset for missing-spectral vehicle ReID.
When a single spectrum is missing, TOP-ReID consistently outperforms DENet, achieving a remarkable mAP increase of over 8\%.
In scenarios where both NIR and TIR spectra are absent, our model demonstrates substantial improvements, with a mAP increase of 5.3\% and a Rank-1 improvement of 3.6\%.
In comparison to DENet, our model excels in handling a wider range of missing scenarios, affirming the versatility of our framework for object ReID.
\begin{table*}[!h]
    \centering
    \renewcommand\arraystretch{1.12}
    \setlength\tabcolsep{5pt}
    \begin{tabular}{cc|cccccccccccc}
    \hline
    \multicolumn{2}{c|}{\multirow{2}{*}{\textbf{Methods}}} &  \multicolumn{2}{c}{\textbf{M (RGB)}} & \multicolumn{2}{c}{\textbf{M (NIR)}} & \multicolumn{2}{c}{\textbf{M (TIR)}} & \multicolumn{2}{c}{\textbf{M (RGB+NIR)}} & \multicolumn{2}{c}{\textbf{M (RGB+TIR)}} & \multicolumn{2}{c}{\textbf{M (NIR+TIR)}} \\
    \cline{3-14}
    & & \textbf{mAP} & \textbf{R-1}  & \textbf{mAP} & \textbf{R-1} & \textbf{mAP} & \textbf{R-1} & \textbf{mAP} & \textbf{R-1} & \textbf{mAP} & \textbf{R-1} & \textbf{mAP} & \textbf{R-1} \\
     \hline
    \multirow{2}{*}{\textbf{Multi}}
     & DENet  & - & - & \underline{62.0} & \underline{85.5} & \underline{56.0} & \underline{80.9} & - & - & - & - & \underline{50.1} & \underline{74.2} \\
     & \textbf{TOP-ReID} & \textbf{70.6} & \textbf{90.6} & \textbf{77.9} & \textbf{94.5}& \textbf{64.0} & \textbf{81.5}  & \textbf{42.5} & \textbf{69.3} & \textbf{45.9} & \textbf{65.4} & \textbf{55.4} & \textbf{77.8} \\
     \hline
    \end{tabular}
    \vspace{-2mm}
    \caption{Experimental results of missing-spectral tasks on RGBNT100. “M (X)" stands for missing the X image spectra.}
    \label{tab:missing-spectral vehicle ReID}
    \vspace{-0mm}
\end{table*}
\begin{figure*}[h]
    \centering
    \includegraphics[width=0.88\textwidth]{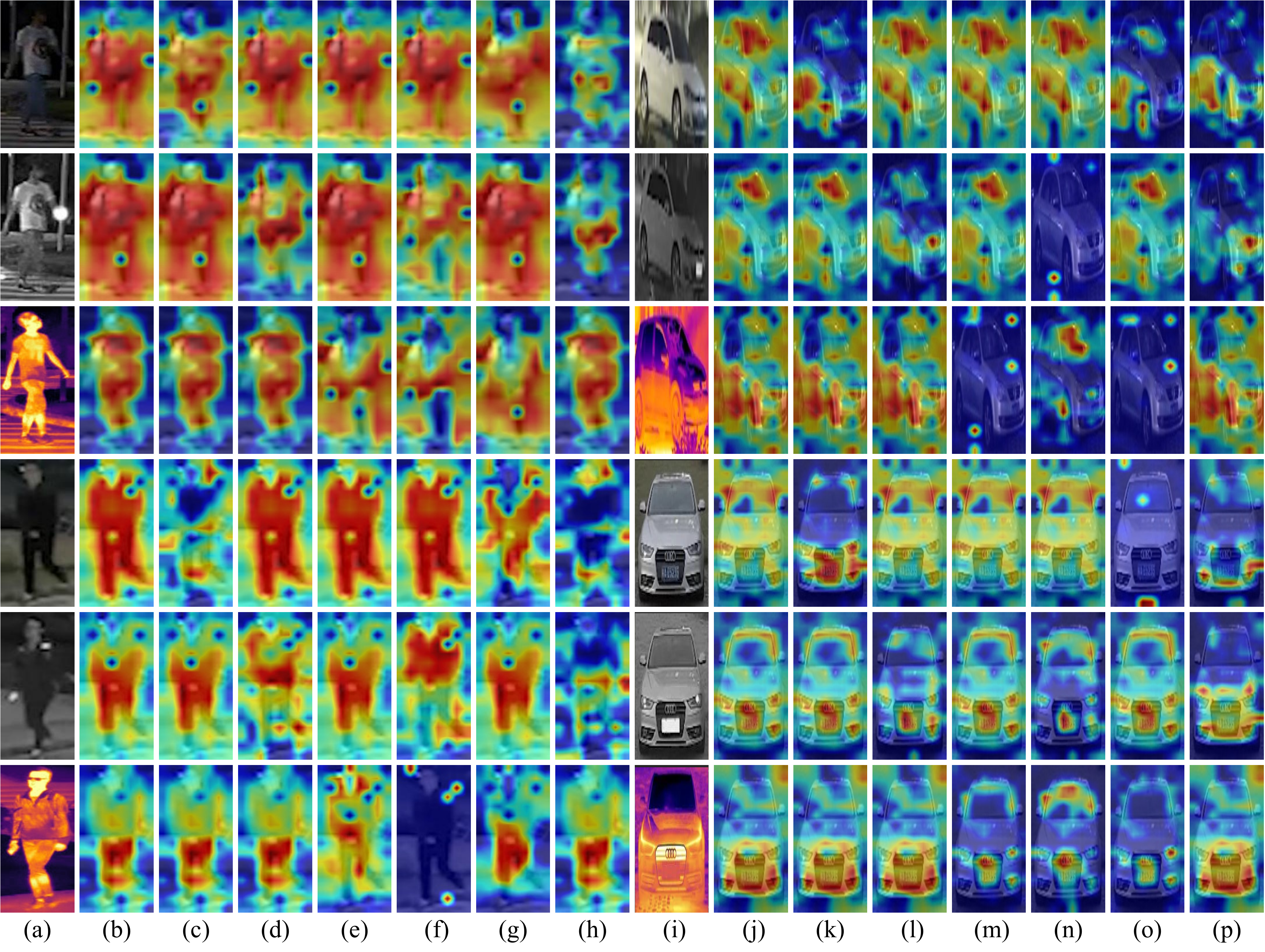}
    \caption{Discriminative attention maps. (a) Input person images; (b) Full; (c) M (RGB); (d) M (NIR); (e) M (TIR); (f) M (NIR+TIR); (g) M (RGB+TIR); (h) M (RGB+NIR).
    (i) to (p), follow the same setting with the vehicle inputs.}
    \label{appendix:grad-cam}
    \end{figure*}
\section{C Visualizations}
\subsection{C.1 Token alignments in TPM.}
In Fig. \ref{fig:align}, we present the cosine similarity distribution between the class tokens and the patch tokens from other spectra.
As additional evidences, one can find that the cosine similarities are higher after TPM, indicating better alignments.
\subsection{C.2 Discriminative Attention Maps.}
As shown in Fig. \ref{appendix:grad-cam}, we visualize the attention maps of instances on RGBNT201 and RGBNT100.
Panel (b) demonstrates the attention regions of each branch when no spectral is missing.
It can be observed that there are certain differences for different spectra, especially for TIR spectrum, which clearly separates the person from the background.
Meanwhile, all branch can capture the discriminative regions of person.
Panels (c), (d), and (e) represent the effects of using the other two spectra to reconstruct the missing spectral.
It can be seen that the model restores discriminative attention regions.
Panels (f), (g), and (h) represent the reconstruction of the other two spectra using only the existing spectrum.
It is noticeable that the model's reconstruction ability becomes weak, but it is still able to introduce regions that the existing spectral did not focus on.
These visualizations illustrate the effectiveness of TOP-ReID.
In addition, we provide attention maps for the vehicle inputs.
Just like the (i) to (p) cases, which share the same settings as the person inputs, we can observe that different spectra focus on distinct regions of the vehicles.
Moreover, the model demonstrates effective reconstruction, particularly when one spectrum is missing.
To sum up, these visualizations provide substantial evidence of the effectiveness of TOP-ReID.
\subsection{C.3 Reconstructed Tokens in CRM.}
\begin{figure*}[t]
        \centering
        \includegraphics[width=0.9\textwidth]{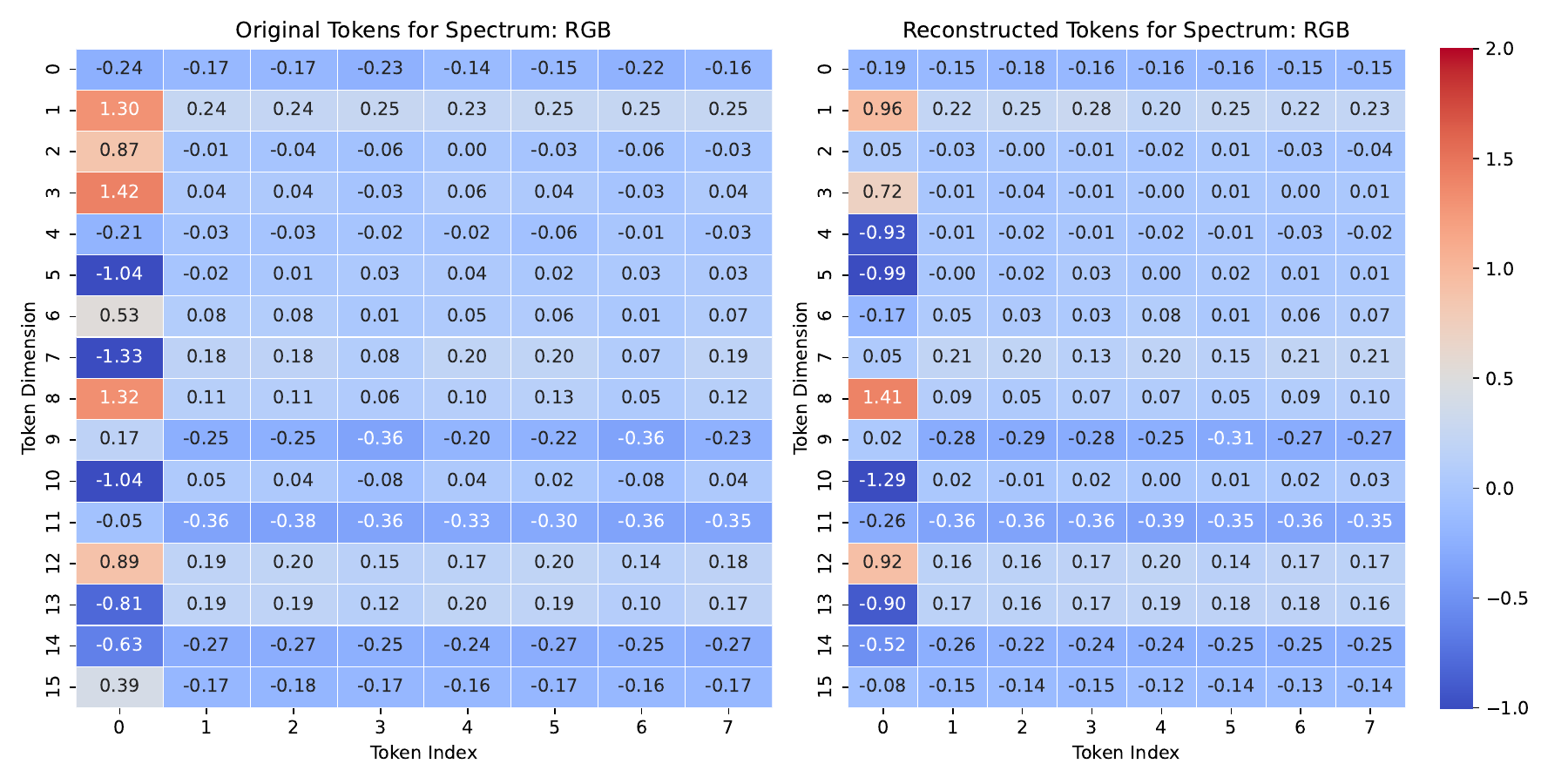}
        \includegraphics[width=0.9\textwidth]{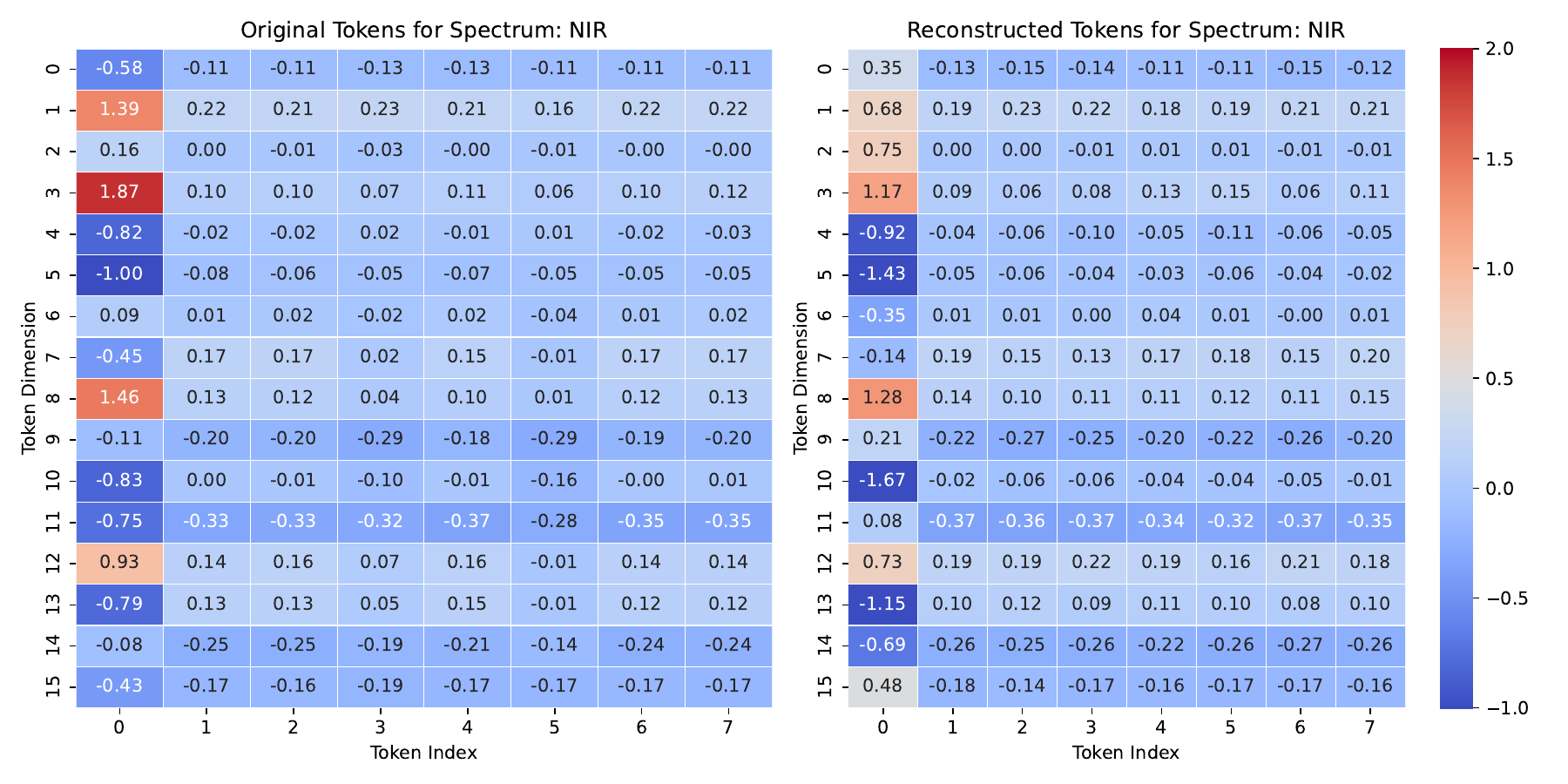}
        \includegraphics[width=0.9\textwidth]{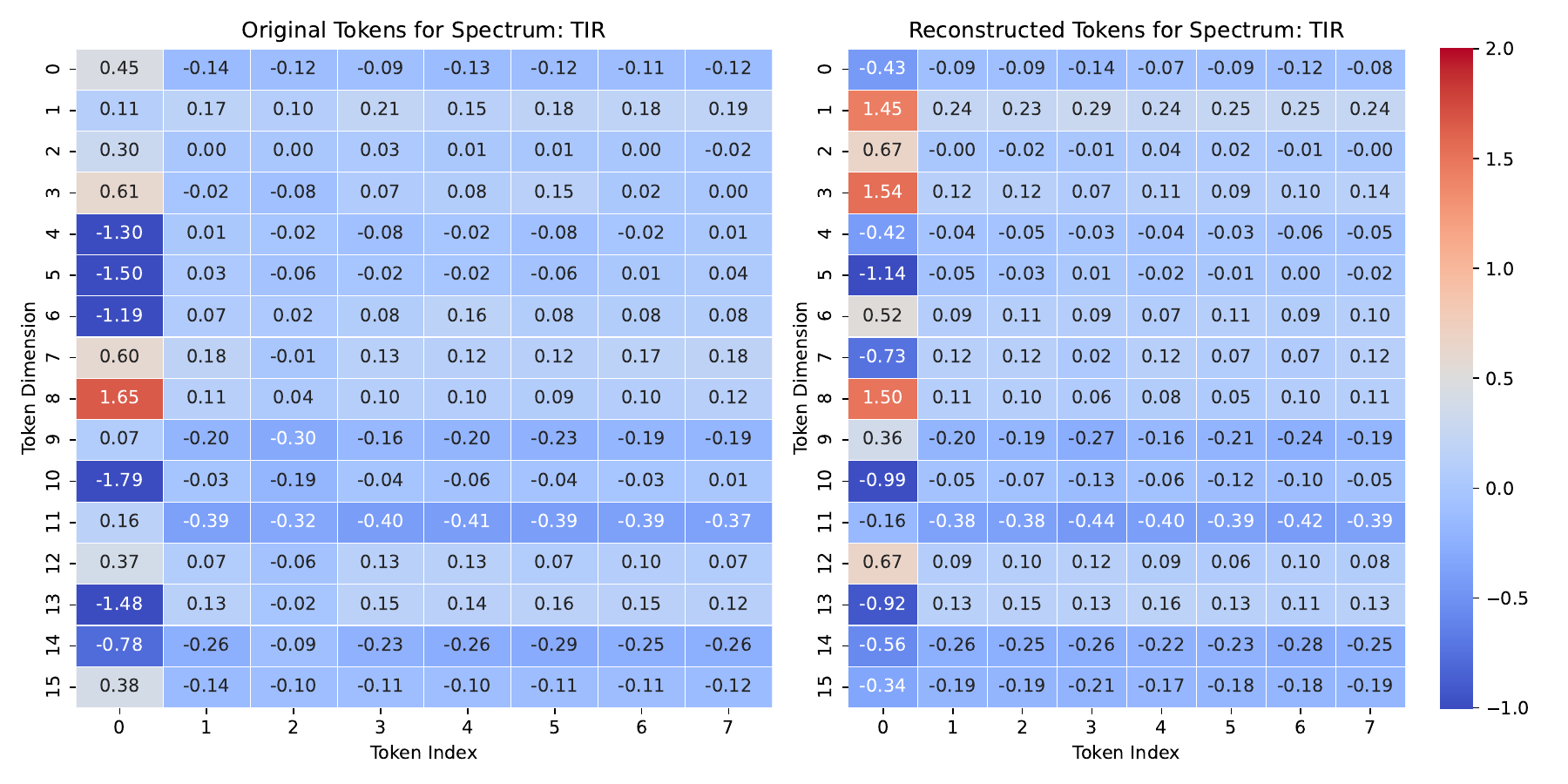}
        \caption{Comparison between the original tokens and the reconstructed tokens.}
    \label{appendix:crm}
\end{figure*}

We further visualize the comparison between the reconstructed tokens and original tokens in CRM.
As shown in Fig. \ref{appendix:crm}, from top to bottom, there are schematic representations of the matrix values for the reconstructed RGB, NIR, and TIR spectrum.
To clearly illustrate the visualization, we use the first 8 tokens, including the class token, and the first 16 embedding dimensions for visualization.
In terms of numerical distribution, the reconstructed tokens closely approximate the original tokens, especially for the distribution of patch tokens.
However, concerning the cls token, although there is some numerical difference, its importance level remains consistent with the original token.
Therefore, through the CRM, we achieve effective alignment with original tokens, thereby assisting the missing-spectral scenarios.

\end{document}